\documentclass[10pt,twocolumn,letterpaper]{article}

\usepackage[pagenumbers]{wacv}

\usepackage{graphicx}
\usepackage{amsmath}
\usepackage{amssymb}
\usepackage{booktabs}

\usepackage{times}
\usepackage{epsfig}
\usepackage{comment}
\usepackage{multirow}
\usepackage{array}
\usepackage[toc,page]{appendix}
\newcommand{\specialcell}[2][c]{\begin{tabular}[#1]{@{}c@{}}#2\end{tabular}}
\usepackage{pifont}
\usepackage{textcomp}

\usepackage[pagebackref,breaklinks,colorlinks]{hyperref}

\usepackage[capitalize]{cleveref}
\crefname{section}{Sec.}{Secs.}
\Crefname{section}{Section}{Sections}
\Crefname{table}{Table}{Tables}
\crefname{table}{Tab.}{Tabs.}

\begin{document}

\title{Subjective Face Transform using Human First Impressions}

\author{Chaitanya Roygaga\(^1\), Joshua Krinsky\(^1\), Kai Zhang\(^2\), Kenny Kwok\(^1\), Aparna Bharati\(^1\)\\
\(^1\)Lehigh University, \(^2\) University of Notre Dame\\
\(^1\)Bethlehem, PA, USA, \(^2\)South Bend, IN, USA\\}

\maketitle

\begin{abstract}
   Humans tend to form quick subjective first impressions of non-physical attributes when seeing someone's face, such as perceived trustworthiness or attractiveness. To understand what variations in a face lead to different subjective impressions, this work uses generative models to find semantically meaningful edits to a face image that change perceived attributes. Unlike prior work that relied on statistical manipulation in feature space, our end-to-end framework considers trade-offs between preserving identity and changing perceptual attributes. It maps latent space directions to changes in attribute scores, enabling a perceptually significant identity-preserving transformation of any input face along an attribute axis according to a target change. We train on real and synthetic faces, evaluate for in-domain and out-of-domain images using predictive models and human ratings, demonstrating the generalizability of our approach. Ultimately, such a framework can be used to understand and explain trends and biases in subjective interpretation of faces that are not dependent on the subject's identity. This is demonstrated with improved model performance for first impression prediction when augmenting the training data with images generated by the proposed approach for a wider range of input to learn associations between face features and subjective attributes. 
\end{abstract}
\vspace{-0.65cm}

\section{Introduction}

\begin{figure}[htbp]
    \centering
    \includegraphics[width=7.5cm]{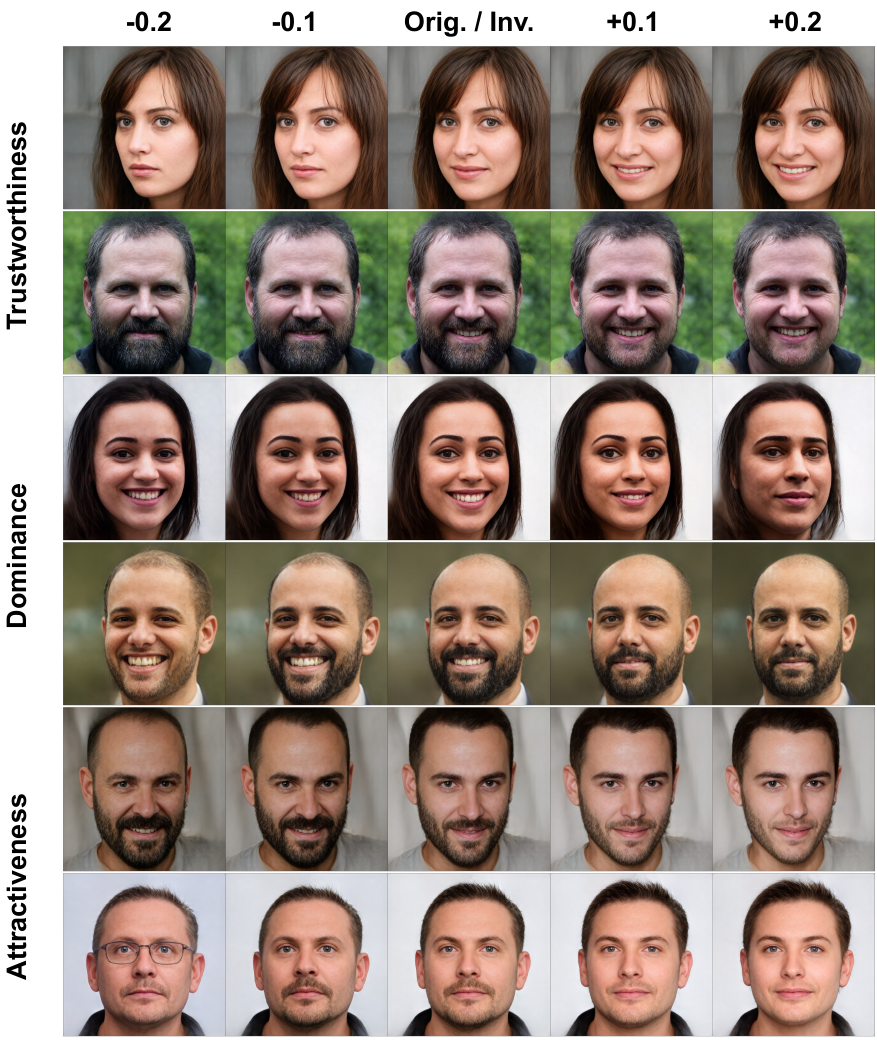}
    \vspace{-0.2cm}
    \caption{
    We propose a method of exploring possible identity-disentangled semantic variations of a face image in a latent space that encodes subjective attributes. Upon learning the mapping between the latent representations of images and the corresponding subjective attribute scores, any image can be transformed based on a desired score change for a particular attribute -- with all possible score-based variations scaled between $[0,1]$. \textit{Original (Orig.) / Inversion (Inv.)} shows the original image as reconstructed using inversion of the latent representation learned by the GAN~\cite{karras2020training}.}
    \label{fig:teaser_face_transform}
    \vspace{-0.6cm}
\end{figure}

Face perception skills are not only exemplified through competence in tasks such as face detection and recognition but also, more abstract inferences of complex emotions and first impressions related to traits such as trustworthiness and attractiveness~\cite{keating1981culture, schiller2009neural, peng2024isfb}. Although not always accurate~\cite{zebrowitz2017first}, subjective face judgments are an integral part of day-to-day interpersonal communication, with first impressions getting more reliable over time~\cite{capozzi2020attention}. Some individuals may find it challenging to interpret certain facial expressions and form accurate first impressions~\cite{bekele2013understanding}. This difficulty can result in negative interactions, including misunderstandings, fear, deception, and, in some cases, rejection. Understanding aspects that could improve the experience during interpersonal interactions can be extremely useful, specially, after a pandemic-induced hiatus in practicing social interactions. First impressions from faces are even more important for online interactions, as other sources such as body language and a full range of continuous stimuli can be absent~\cite{cortez2017first, hu2018first}.

Invariant aspects of a face i.e., \emph{identity}~\cite{wang2021deep} and its covariates i.e., \emph{expression, pose, gaze direction} contribute to a perceived first impression~\cite{haxby2000distributed}. Facial features that yield high and low attractiveness or trustworthiness 
have mostly been studied for a small set of stimuli with different identities (\emph{inter-identity subjectivity})~\cite{little2011facial}. However, non-identity features can also lead to variable perceived judgments for the same person's face 
(\emph{intra-identity subjectivity}). Understanding properties of faces that result in a specific subjective first impression can not only help generate a better curated set for psychology-based debiasing experiments~\cite{jaeger2020can}, by capturing more variations, but also analyze possible first impression responses for a face. 
To understand variations in a face image leading to different
subjective impressions by humans, and inspired by generative models that conditionally edit an original image with high fidelity and control~\cite{hu2022style, wang2024maniclip, suwala2024face},
we \textit{propose a Continuous Normalizing Flow-based technique that uses GAN latent space to generate variants of a given face that correlate with perceived first impressions} (Fig.~\ref{fig:teaser_face_transform}). The proposed framework maps subjective attributes to the latent space of editable face features and learns attribute-relevant transforms that largely retain the original identity of the face, in the context of face recognition~\cite{wang2021deep}. The method achieves smooth image modifications without explicit edit directions~\cite{shen2020interpreting, peterson2022omi}. Our experiments compare results from training on real images from the FFHQ~\cite{karras2019style} and CelebAMask-HQ~\cite{CelebAMask-HQ} datasets, and synthetic images generated using StyleGAN2-ADA~\cite{karras2020training}. To demonstrate the generalizability of our approach, trained models are evaluated on four out-of-domain datasets - Chicago Face~\cite{ma2015chicago}, SCUT-FBP5500~\cite{liang2018scut}, US Adult Faces~\cite{bainbridge2013intrinsic}, and One Million Impressions (OMI)~\cite{peterson2022omi}. 

We further demonstrate the application of synthetically generated face variations using our proposed pipeline to improve models for predicting perceived first impressions. First impression prediction models are based on crowd-sourced opinions tied to subjective attribute labels from face stimuli. These models may misinterpret impressions as personality traits instead of learning attribute-specific facial features independent of identity, which can introduce bias. The bias gets exacerbated when there aren't enough examples of a specific subject for the model to learn from.  Many existing datasets feature only one face per identity~\cite{peterson2022omi, ma2015chicago, liang2018scut, bainbridge2013intrinsic,  mccurrie2018convolutional}. Synthetic data can help reduce the bias due to limited training data diversity~\cite{jaipuria2020deflating, zhang2024relative}. We propose training impression prediction models with synthetic face variations across score distributions for better generalization and attribute-specific predictions. This enables the model to learn a broader attribute-based feature space for the original face stimuli identities.

The contributions of this work can be summarized as:

\begin{enumerate}
\itemsep0em
    \item Presenting a novel flow-based framework that transforms faces according to user-desired change in the perceived first impression attribute score. The proposed solution generates perceptual variants --- based on attributes like trustworthiness, dominance, and attractiveness --- largely preserving 
    facial features relevant for face recognition using a novel identity loss.
    \item Conducted extensive evaluations showing edits to be consistent with human perception and comparing results from training using real and synthetic faces.
    \item Visualizing objective edits, performed by the pipeline, highlighting the biases from human annotators participating in the first impression tasks.
    \item Demonstrating that fine-tuning with synthetic faces generated using the pipeline improves the performance and out-of-domain generalizability of facial first impression prediction models.
\end{enumerate}

\section{Relevant Literature}

For the editing of face images, previous work has dealt with objective semantic attributes such as pose, illumination, facial hair, 
and smile~\cite{shen2020interpreting, karaouglu2021self, yang2021l2m, yang2022s2fgan, hou2022guidedstyle, wei2022hairclip, mohammadbagheri2024identity}. \textit{Latent space editing of face images based on subjective attributes} is relatively less explored, and based on perceptual similarity in the face recognition context~\cite{wang2021deep}, these methods do not preserve identity~\cite{saquil2018ranking, diamant2019beholder, yang2021novel, peterson2022omi}. Existing work related to the proposed approach are discussed under two broad categories --- face image editing based on constraints and predicting subjective attributes relevant to facial first impressions.

\subsection{Attribute-based Face Editing}

\subsubsection{Objective Attribute Editing}

\textbf{GAN-based methods -} Existing approaches adopt two techniques for transforming face images based on learned training data distribution --- conditional image generation and latent space manipulation. Conditional image generation takes the original image as input and transforms it based on a given conditional label. To retain identity, conditional generative networks such as FaceFeat-GAN~\cite{shen2018facefeat}, FaceID-GAN~\cite{shen2018faceid}, and others~\cite{bao2018towards} use an auxiliary network. Although the quality of generated face images is decent, such models lack control over explicit face features.

\begin{figure*}[t]
    \centering
    \includegraphics[width=18cm]{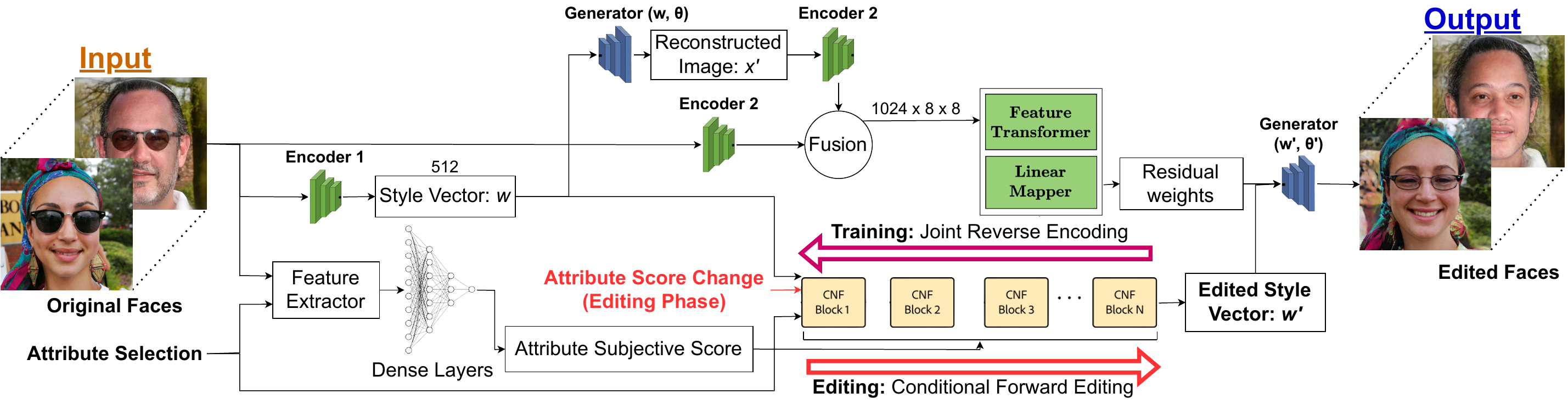}
    \vspace{-0.2cm}
    \caption{Proposed face transformation method. Original face image is given as input, and the first phase involves extracting image features (image latents) and the corresponding human-like score predictions, for the selected attribute. Then, to generate disentangled and continuous mapping of image latent features with the corresponding scores, image latents are continuously evolved according to predicted attribute scores. The editing phase uses the mapping to directly transform the input image latent $w$ according to the desired score change. Finally, inversion step reconstructs the transformed image from edited latent $w'$.} 
    \label{fig:face_transform_pipeline}
    \vspace{-0.5cm}
\end{figure*}
 
GAN latent spaces encode similarities and variations between 
image features in a compressed representation which allows discovering edit directions in the latent space that correspond to semantic changes in image space~\cite{hu2022style}. Different linear~\cite{tov2021designing, lin2021attribute, parmar2022spatially, wang2022high, katsumata2024revisiting} and non-linear approaches~\cite{yang2022s2fgan, hou2023deep} have been proposed to achieve controllable disentangled latent spaces. InterFaceGAN~\cite{shen2020interpreting, shen2020interfacegan} was the first to find linear latent spaces that correspond to meaningful face transformations, allowing for additional control over generated images for attributes like pose. GANSpace~\cite{harkonen2020ganspace} introduced linear control over GAN latent spaces using PCA to uncover interpretable directions for modifications like gender and hair color. Subsequent works, such as Hou~\etal~\cite{hou2022guidedstyle} and VecGAN++~\cite{dalva2023image}, enhanced latent space editability by employing attention networks to refine style layer selection and learn attribute-specific editing directions.
Alternatively, global manipulation directions assigned using target attributes in input text prompts has been shown to improve objective attribute editing in the StyleGAN2~\cite{karras2020analyzing} space~\cite{kocasari2022stylemc}. However, these methods rely on manual annotations in the form of such target attributes. Xu~\etal~\cite{xu2022predict} reduce reliance on manual annotations by using a pretrained CLIP~\cite{radford2021learning} model to predict related attributes from input queries. Their manipulation model~\cite{patashnik2021styleclip} enforces disentanglement by penalizing transformations of entangled attributes while editing multiple attributes simultaneously.

Although these methods have improved the disentanglement of the editing feature space for higher original context retention and smoother edits towards target attributes, they depend on the ability of prior inversion methods for real image editing. Cao~\etal~\cite{cao2024decreases} simultaneously improve inversion and editing capabilities using Domain-Specific Hybrid Refinement (DHR). 
For each training image, in-domain regions update the generator's weights, while out-domain features are directly inverted to preserve spatial context, with DHR combining both for final edits. Xu~\etal~\cite{xu2022transeditor} introduce a transformer-based architecture to disentangle style and content spaces, using cross-attention to re-weight style with content embeddings, enabling context-preserving edits.

Non-linear latent space editing for attributes like head rotation and face shape is performed by frameworks such as StyleRig~\cite{tewari2020stylerig} and StyleFlow~\cite{abdal2021styleflow}. The former uses a 3D morphable model to generate face meshes for the desired attribute change in the StyleGAN~\cite{karras2019style} latent space, by proposing semantic control parameters for those directions. Due to the dependency on specific predefined control parameters, its application for subjective attribute mapping is limited. The latter method uses Continuous Normalizing Flow (CNF) to map latent combination of transformations resulting in overall changes towards a target attribute. To improve user control over the edited objective attributes, StyleCLIP~\cite{patashnik2021styleclip}, ManiCLIP~\cite{wang2024maniclip}, and AnyFace++~\cite{sun2024anyface++} provide a \textit{text-based} interface that translates user input to non-linear edits in StyleGAN's~\cite{karras2019style} latent space, an alternative form of mapping when compared to~\cite{abdal2021styleflow}. ManiCLIP~\cite{wang2024maniclip} in particular combines CLIP embeddings from target attributes of similar difficulty during training, relying on the ease (\textit{e.g.} lipstick) and difficulty (\textit{e.g.} chubby) in editing certain attributes. This allows the model to predict a combination offset $\triangle$\textit{w} for the original latent \textit{w} to transform the image. Recent methods have also focused on improving the identity retention during face editing, like PluGeN4Faces~\cite{suwala2024face}, Arc2Face~\cite{papantoniou2024arc2face}, and DreamIdentity~\cite{chen2024dreamidentity}, while others have proposed methods to output face edits of variable resolution~\cite{yang2023styleganex}.

\textbf{Diffusion-based methods -} Diffusion models allow for editing directions similar to GANs~\cite{yu2023freedom, dalva2024noiseclr} --- FaceComposer~\cite{wang2024facecomposer} utilizes latent diffusion model~\cite{rombach2022high} and pre-extracted identity features from ArcFace~\cite{deng2019arcface} to transform faces across target attributes like hair. PCA has also proven to be useful in discovering interpretable edits for attributes like glasses, pose, and smile~\cite{haas2024discovering}. Diffusion autoencoders~\cite{preechakul2022diffusion} provide another alternative for edits in the diffusion latent space. Lu~\etal~\cite{lu2024hierarchical} suggest that the attribute entanglement can be reduced using a hierarchical structure. Instead of using a single 512-dimensional feature from the final layer, they extract feature maps from the lower and middle layers of the semantic encoder, preserving image details like color and texture from low-level features and the overall structure from high-level features. 

Extending the idea of a non-linear, disentangled, and editable latent space, Hu~\etal~\cite{hu2024latent} use a transformer-based U-ViT to identify interpretable semantic directions in the pretrained autoencoder latent \textit{u}-space. Edits are then performed in the sampling (backward) step in the proposed \textit{u}-space, guided by the target text query. These methods perform context-retaining face edits over objective attributes, but require additional steps during inference to perform these edits. To reduce the computational requirements during inference, LEdits++~\cite{brack2024ledits++} introduces the idea of semantically grounding the target editing regions using binary masks for each edit instruction, allowing for disentangled edits.

\subsubsection{Subjective Attribute Editing} 

The previously discussed methods provide identity-disentangled transformations of face images over user-defined attribute directions, but they are limited to objective-style edits over attributes like hair, smile, skin color, and expression. The fine-grained editing controls demonstrated by them motivates us to \textit{explore such transformations of face images over subjective attributes instead}, like Trustworthiness and Dominance. Such a mapping is non-trivial to learn. The first impression of subjective attributes -- personality traits that we assign to someone upon perceiving their face for the first time~\cite{zebrowitz2017first} -- are instinctive~\cite{todorov2005inferences, todorov2009evaluating, ritchie2017forming} and highly dependent on facial characteristics~\cite{olivola2010fooled, yan2015your, rojas2011automatic, bhat2019first}. The impression data primarily represents a small group of annotators, which allows room for differences in opinion~\cite{zebrowitz2017first}. Absence of a factual ground truth makes calibrating the \textit{degree of change of a face image expected by a human} difficult.

Early work such as
~\cite{saquil2018ranking, yang2021novel} introduced editability of faces over specific subjective attributes -- \textit{masculinity} and \textit{age}, but generate lower quality faces. Diamant \etal~\cite{diamant2019beholder} introduce a PGGAN~\cite{karras2017progressive} variant to transform ($\uparrow$) faces over \textit{attractiveness}. IricGAN~\cite{ning2022face} use a novel hierarchical feature combination (HFC) module to extract and filter multi-scale features from the input samples, which are then combined with an attribute regression module (ARM) that introduces the disentanglement and intensity of edits to be performed over the target attributes. Their results show identity loss 
over the edits.

Peterson~\etal~\cite{peterson2022omi} explore subjective style face transformations that vary along perceived attribute directions. They edit images with respect to user desired change using simple vector arithmetic and a mapping between input images and linear models trained to capture human perception of the target attributes. To our knowledge, \cite{peterson2022omi} is currently the \textit{only work} that performs face transformations over $\downarrow$ and $\uparrow$ subjective attribute directions and aligns with our goal of developing a framework that models a complete spectrum of possible transformations over a variety of subjective attributes. While their transformations are smooth with visible edits, they show perceptually significant changes in identity (Table~\ref{table:omi_comparison}). Our inversion step~\cite{dinh2022hyperinverter}, along with a novel identity loss, allows us to 
freely use the editable latent space $w$ for transformation while retaining perceptually significant identity features. Due to their ability to perform well even with less training data~\cite{zhang2024improving} and narrow distributions~\cite{melnik2024face, hou2024augmentation}, we use StyleGAN2-ADA~\cite{karras2020training} in our flow-based pipeline. The training method (ADA) of~\cite{karras2020training} is optimized to train a diverse generator model with a few thousand samples --- after performing our image filtering steps, we are left with 10883 training images\footnote{Please refer Appendix~\ref{section:attribute_and_image_data}.}. GANs are also computationally efficient during inference when compared to diffusion models, and can generate samples during inference without additional steps.

\subsection{Predictive Models for Subjective Judgments}

Different model architectures 
have been trained for predicting first impressions from faces. McCurrie~\etal~\cite{mccurrie2018convolutional} train a CNN-based regression framework with 6300 images sampled from the AFLW~\cite{koestinger2011annotated} dataset, annotated using a crowd-sourcing platform. With a higher variance during scoring for subjective attributes like dominance and trust, their models yield a performance of 0.51 and 0.43 $R^2$, respectively. Our pipeline considers their training approach as a baseline to pre-train the first impression prediction models. 

To understand the feature combinations that humans observe to form first impressions, the authors in~\cite{mccurrie2018convolutional} use activation maps~\cite{selvaraju2017grad} and study the importance of facial areas in rating these qualities. For example, trustworthiness and dominance tend to rely more on the mouth and chin areas. Such studies help in understanding the face features that evoke specific responses with respect to an attribute and inspire exploration of corresponding edits in the transformation space of face image editing pipelines. Similarly, Messer~\etal~\cite{messer2019predicting} trained CNN models for feature extraction and binary classifiers to detect human-like impressions of warmth and competence from input face images. They discussed that visual stimuli in the form of faces elicit varying levels of first impressions over different subjective attributes, and it is vital to study the complete spectrum of stimuli-to-score mappings over such variations. Our pipeline provides one of the first methods in facilitating a synthetically generated set of visual stimuli that reflect such identity-disentangled semantic variations of a face image in a latent subjective attribute feature space.

\section{Proposed Method}
\label{section:proposed_method}

We propose a conditional face transformation pipeline, capable of learning continuous mapping of face features to their corresponding ratings of subjective attributes such as Trustworthiness --- allowing semantic-aware exploration of identity-disentangled variations. The mapping can later be employed to transform a face along a spectrum of the desired subjective attribute changes. Our approach (Figure~\ref{fig:face_transform_pipeline}) consists of three phases; (a) extracting face latent $w$ and corresponding human-like perceived attribute score, (b) mapping latents and scores, with corresponding latent transformation in the editing phase, and (c) image inversion process to output the transformed image.

\subsection{Image Latent and Subjective Score Extraction}
\label{section:latent_score_extraction}

The first phase of the pipeline performs two tasks --- the transformation of input face images into an editable feature space, and the prediction of human-like consensus scores for the selected subjective trait via a trained model.

\noindent\textbf{Projecting face images onto latent space - } To enable the framework to edit any given image (real or synthetic), we perform GAN inversion. Inversion primarily depends on the generator architecture and the type of latent space on which the input image is projected, and is expected to recover almost true reconstruction --- $w$ space provides higher editability~\cite{tov2021designing}. For generating an editable image feature vector from the original image, the encoder ($E_1$) takes as input the original image (x) and projects it onto the $w$ latent space, $w_{512} = E_1(x)$. $w_{512}$ is then input to the mapping module for feature transformation, with contextual mapping to a target attribute's subjective score. Simultaneously, encoder ($E_2$) generates a higher dimensional latent code from the input image, corresponding to face-specific details relevant to the GAN inversion space. The downstream fusion (layer-wise concatenation) of embeddings extracted from the original and reconstructed images (Fig.~\ref{fig:face_transform_pipeline}) gives us an \textit{appearance code}, $h_{1024x8x8}=E_2(x') \oplus E_2(x)$, allowing identity-specific reconstruction. Encoders 1 \& 2 use ResNet-50~\cite{he2016deep} as backbone.
We discuss image inversion details in Appendix~\ref{section:image_inversion}.

\noindent\textbf{Subjective attribute score extraction - } Models trained to predict scores for subjective attributes can emulate human-like perspective in the absence of humans and can be time efficient. They output the original attribute score, \textit{i.e.} how trustworthy, dominant, or attractive a face image looks, which can be compared to the score of the transformed image to give us evidence of the change in human perception. We train separate models for each attribute, with CNN-based architectures as proposed in~\cite{mccurrie2018convolutional} for regression-based prediction of the subjective attributes (Appendix~\ref{section:subjective_attribute_prediction}).

\subsection{Mapping and Editing Subjective Attributes}
\label{section:mapping_module}

The second phase involves mapping image latent vectors to their corresponding subjective scores to get identity-disentangled variations in the semantic feature space of the selected subjective attribute. The mapping allows for possible face transformations in that feature space according to a user-desired value of change --- variations are scaled such that 0 corresponds to the least possible trustworthiness, for example, and 1 corresponds to the most trustworthiness an image can elicit. We employ the base architecture of~\cite{abdal2021styleflow} which builds on top of FFJORD~\cite{grathwohl2018ffjord}. While~\cite{abdal2021styleflow} edits face attributes based on objective features such as image illumination and face pose, we provide face transformation mappings with subjective attribute scores (like Trustworthiness) as targets. Our training losses consider transformation of the latent $w$ with respect to the target score of a given attribute --- intuitive to a human user --- with CNF blocks learning the combined representation of multiple objective edits (age, facial hair, etc.) corresponding to a given score.

The basic unit in FFJORD~\cite{grathwohl2018ffjord} is the Continuous Normalizing Flow (CNF) block~\cite{chen2018neural}. It is a sequence of latent transformations that map unknown to known distributions; unknown distributions are the non-linear feature space of subjective attributes. Our mapping task takes as input the $w$ latent vector and context (\textit{$S_t$+}), provided to all CNF blocks in stack (Fig.~\ref{fig:face_transform_pipeline}). Two kinds of condition information are required in the \textit{mapping context}: the time variable \textit{t} and the desired attribute's score \textit{$S_t$}. We transform~\cite{abdal2021styleflow} \textit{t} with a broadcast operation \textit{B} to match spatial dimensions of \textit{$S_t$}, later concatenating \textit{t} with \textit{$S_t$} to form \textit{$S_t$+}. \textit{$S_t$+} represents input subjective scores with respect to the time variable \textit{t}.

The first step of the CNF Block is represented as a differential equation where $w$ is the input distribution of image features, and mapping between $w$ and $S_t$+ allows training of \textit{selected} neural network's parameters $\theta$. To model function $\phi$, we employ a series of ConcatSquash gate-bias modulation~\cite{grathwohl2018ffjord, yang2019pointflow}. This module works as a linear transformation on the latent code, and the weight of this transformation is learned from the context of the subjective attribute score --- from the transformation of context \textit{$S_t$+}. Eqns.~\ref{eqn:flow_loss_2} and~\ref{eqn:flow_loss_3} represent the loss calculated on the transformation of $w$.

\small
\begin{equation}
w(t_1) = w(t_0) + \int_{t_0}^{t_1} \phi(w(t),S_t+;\theta) \,dt
\label{eqn:flow_loss_2}
\end{equation}

\begin{equation}
\log p(w(t_1)) = \log p(w(t_0)) + \int_{t_0}^{t_1} T_r({\frac{\partial\phi}{\partial w(t)}}) \,dt
\label{eqn:flow_loss_3}
\end{equation}
\normalsize

\begin{table*}[htbp]
\begin{center}
\begin{small}
\caption{Setup: \textit{No ID Loss}. Identity Similarity (IS), Perceptual Distance (PD), and Fréchet Inception Distance (FID) are used for evaluation of proposed conditional editing pipeline. IS=1 shows perfect identity retention, and FID=0 for most realistic quality. $\uparrow$ PD scores imply more perceptually visible edits. Combined interpretation of IS and PD gives us overall editing performance.}
\vspace{-0.3cm}
\label{table:final_results}
\begin{sc}
\renewcommand{\arraystretch}{1.4}
\begin{tabular}{ >{\raggedright\arraybackslash}p{1cm} |
>{\raggedright\arraybackslash}p{0.3cm} |>{\raggedright\arraybackslash}p{2.2cm} |>{\raggedright\arraybackslash}p{0.85cm} |>
{\raggedright\arraybackslash}p{0.85cm} |>{\raggedright\arraybackslash}p{0.85cm} |>{\raggedright\arraybackslash}p{0.85cm} |>{\raggedright\arraybackslash}p{0.85cm} |>{\raggedright\arraybackslash}p{0.85cm} |>{\raggedright\arraybackslash}p{0.85cm} |>{\raggedright\arraybackslash}p{0.85cm} |>{\raggedright\arraybackslash}p{0.85cm}}
 \hline
 \multirow{2}{1cm}{\bfseries Train Set} & \multicolumn{2}{>{\raggedright\arraybackslash}p{1.5cm}|}{\multirow{2}{1.2cm}{\bfseries Evaluation Set}} & \multicolumn{3}{>{\raggedright\arraybackslash}p{3cm}|}{\bfseries Trustworthiness} & \multicolumn{3}{>{\raggedright\arraybackslash}p{3cm}|}{\bfseries Dominance} & \multicolumn{3}{>{\raggedright\arraybackslash}p{3cm}}{\bfseries Attractiveness} \\
 \cline{4-12}
  & \multicolumn{2}{>{\raggedright\arraybackslash}p{1.5cm}|}{} & {\bfseries IS $\uparrow$} & {\bfseries PD $\uparrow$} & {\bfseries FID $\downarrow$} & {\bfseries IS $\uparrow$} & {\bfseries PD $\uparrow$} & {\bfseries FID $\downarrow$} & {\bfseries IS $\uparrow$} & {\bfseries PD $\uparrow$} & {\bfseries FID $\downarrow$} \\
 \hline
 \multirow{6}{1cm}{\rotatebox[origin=c]{90}{\specialcell{FFHQ \\ (Real)}}} & \multicolumn{2}{>{\raggedright\arraybackslash}p{1.5cm}|}{FFHQ~\cite{karras2019style}} & 0.819 & 0.253 & \phantom{0}3.20 & 0.803 & 0.281 & \phantom{0}4.68 & 0.853 & 0.247 & \phantom{0}3.24 \\
  & \multicolumn{2}{>{\raggedright\arraybackslash}p{1.5cm}|}{Synthetic} & 0.800 & 0.253 & \phantom{0}3.06 & 0.783 & 0.280 & \phantom{0}4.77 & 0.841 & 0.243 & \phantom{0}2.98 \\
 \cline{2-12}
  & \multirow{4}{0.3cm}{\rotatebox[origin=c]{90}{Out-Domain}} & CFD~\cite{ma2015chicago, ma2021chicago} & 0.794 & 0.267 & 15.47 & 0.876 & 0.230 & 10.02 & 0.883 & 0.222 & \phantom{0}9.19 \\
  &  & SCUT~\cite{liang2018scut} & 0.802 & 0.261 & 13.68 & 0.824 & 0.255 & 14.68 & 0.871 & 0.232 & 12.06 \\
  &  & UAF~\cite{bainbridge2013intrinsic} & 0.801 & 0.250 & 12.68 & 0.816 & 0.253 & 13.38 & 0.838 & 0.246 & 12.72 \\
  &  & OMI~\cite{peterson2022omi} & 0.805 & 0.235 & 11.43 & 0.790 & 0.260 & 14.75 & 0.851 & 0.222 & \phantom{0}9.97 \\
 \hline
 \multirow{6}{1cm}{\rotatebox[origin=c]{90}{\specialcell{SG2-ADA gen. \\ (Synthetic)}}} & \multicolumn{2}{>{\raggedright\arraybackslash}p{1.5cm}|}{FFHQ~\cite{karras2019style}} & 0.855 & 0.230 & \phantom{0}3.68 & 0.795 & 0.297 & \phantom{0}5.56 & 0.769 & 0.308 & \phantom{0}4.45 \\
  & \multicolumn{2}{>{\raggedright\arraybackslash}p{1.5cm}|}{Synthetic} & 0.842 & 0.226 & \phantom{0}3.53 & 0.779 & 0.292 & \phantom{0}5.68 & 0.751 & 0.303 & \phantom{0}4.27 \\
 \cline{2-12}
  & \multirow{4}{0.3cm}{\rotatebox[origin=c]{90}{Out-Domain}} & CFD~\cite{ma2015chicago, ma2021chicago} & 0.823 & 0.267 & 18.84 & 0.872 & 0.233 & 10.61 & 0.811 & 0.289 & 14.86 \\
  &  & SCUT~\cite{liang2018scut} & 0.842 & 0.243 & 15.66 & 0.828 & 0.259 & 14.75 & 0.795 & 0.288 & 15.23 \\
  &  & UAF~\cite{bainbridge2013intrinsic} & 0.846 & 0.234 & 13.18 & 0.822 & 0.250 & 14.36 & 0.755 & 0.298 & 16.85 \\
  &  & OMI~\cite{peterson2022omi} & 0.848 & 0.211 & 11.26 & 0.786 & 0.276 & 15.81 & 0.772 & 0.274 & 13.36 \\
 \hline
\end{tabular}
\end{sc}
\end{small}
\end{center}
\vspace{-0.5cm}
\end{table*}

Sample inputs are continuously evolved during training according to target context. \textit{Identity} is not inherently preserved during flow-based mapping of face features to corresponding transformed faces --- for a given $z$ vector, different \textit{$S_t$} could result in $w$s that correspond to images with low identity similarity. The amount of identity change induced by the network when \textit{$S_t$} changes is difficult to control without additional identity information. We solve this problem to an extent by using a GAN inversion process with low reconstruction error~\cite{dinh2022hyperinverter}. Results for this vanilla setup are represented under \textit{No ID Loss} (Table~\ref{table:final_results}).

\noindent\textbf{Identity Loss - }We introduce an identity loss, with the idea of 
freezing certain identity-entangled image features while transforming $w$ latent according to the target subjective scores. We first analyze randomly sampled latents from the GAN feature space. To understand the variance in perceptually significant face image features, we consider Principal Component Analysis (PCA) --- our goal is to minimize changes to facial features useful for face matching~\cite{wang2021deep, peng2013efficient, rogers2022roles}. Early principal components are usually associated with coarse features like head orientation and structure~\cite{jozwik2022face}. As faces have a similar basic structure, the ability to separate two identities must be dependent on the more fine-grained features~\cite{andrews2023narrow}. Intuitively, freezing the initial principal components can provide a more fine-grained control over the identity-associated features. As our pipeline performs face edits in the $w$ latent space, we need to first represent the principal components captured from a sufficiently large sample of faces~\cite{andrews2023narrow} within the $w$ feature. Hence, using 300000 randomly sampled $z$, we conduct incremental PCA on the corresponding SG2~\cite{karras2020analyzing} $w$ latents to get first 80 principal editing directions (PED), inspired from~\cite{harkonen2020ganspace}. The first 100 components usually represent 90\%+ of shape and texture variance~\cite{andrews2023narrow}. The $w$ latents are processed as a function of the PED, giving us $w_{80\text{x}512}$ embeddings. We derive latent transformations along two standard deviations (SDs) on either side --- to determine possible transformations along original $w$ latent corresponding to the user-defined transformation spectrum $[-0.2, 0.2]$~(Section~\ref{section:experimental_details}). We get a tensor of shape $w_{80\text{x}5\text{x}512}$ for each initial $w_{512}$. We then take differences of adjacent transformations along the principal components, followed by processing their mean to get $w_{80\text{x}1}$ feature. Freezing few initial dimensions of this feature results in freezing those editing directions in the StyleGAN2 latent space, allowing for a more identity-disentangled face transformation. We experimentally confirm this with initial experiments run for \textit{dominance} by freezing the first \textit{N} PED, where $N = [4, 12, 27]$, discussed further in Section~\ref{section:identity_preservation}.

After experimentally determining possible identity-associated PED, we add an identity regularizer to the original loss while training the mapping model. Similar to the initial steps, we first transform each $w$ as a function of the PED, and calculate transformations along two SDs on either side and take average scores to get $w_{80\text{x}B\text{x}1}$ (B = Batch size). We further \textit{freeze} values associated with first \textit{N} PEDs. This is done for the original $w$ and transformed latents $w'$, after which we calculate L1 Loss between the two tensors. The identity loss is denoted by $\mathcal{L_{\text{reg}}}$, while the original vanilla loss is denoted by $\mathcal{L_{\text{base}}}$. Eqn.~\ref{eqn:final_loss} represents the final loss ---

\small
\begin{equation}
\mathcal{L} = (1-\lambda) * \mathcal{L_{\text{base}}} + \lambda * \mathcal{L_{\text{reg}}}
\label{eqn:final_loss}
\end{equation}
\normalsize
with $\lambda=0.05$, empirically derived. We observe that as the weight ($\lambda$) for the ID Loss term increased, the face transformation ability of the pipeline started to decline, although the original identities were retained. The results of this setup are provided under \textit{ID Loss} (Table~\ref{table:final_results_id_loss}).

\begin{table*}[htbp]
\begin{center}
\begin{small}
\caption{Setup: Comparison with ID loss. Identity Similarity (IS), Perceptual Distance (PD), and Fréchet Inception Distance (FID) are used for evaluation of proposed conditional editing pipeline. IS=1 shows perfect identity retention, and FID=0 for most realistic quality. $\uparrow$ PD scores imply more perceptually visible edits. Combined interpretation of IS and PD gives us overall editing performance. \textit{ID Loss} models are trained by freezing \textbf{4} principal editing directions (PED), as per Table~\ref{table:id_loss_test}. }
\vspace{-0.3cm}
\label{table:final_results_id_loss}
\begin{sc}
\renewcommand{\arraystretch}{1.4}
\begin{tabular}{ >{\raggedright\arraybackslash}p{1cm} | >{\raggedright\arraybackslash}p{0.7cm} |
>{\raggedright\arraybackslash}p{0.3cm} |>{\raggedright\arraybackslash}p{2.2cm} |>{\raggedright\arraybackslash}p{0.85cm} |>
{\raggedright\arraybackslash}p{0.85cm} |>{\raggedright\arraybackslash}p{0.85cm} |>{\raggedright\arraybackslash}p{0.85cm} |>{\raggedright\arraybackslash}p{0.85cm} |>{\raggedright\arraybackslash}p{0.85cm} |>{\raggedright\arraybackslash}p{0.85cm} |>{\raggedright\arraybackslash}p{0.85cm} |>{\raggedright\arraybackslash}p{0.85cm}}
 \hline
 \multirow{2}{1cm}{\bfseries Train Set} & \multirow{2}{0.7cm}{\bfseries ID Loss} & \multicolumn{2}{>{\raggedright\arraybackslash}p{2.1cm}|}{\multirow{2}{1.2cm}{\bfseries Evaluation Set}} & \multicolumn{3}{>{\raggedright\arraybackslash}p{3cm}|}{\bfseries Trustworthiness} & \multicolumn{3}{>{\raggedright\arraybackslash}p{3cm}|}{\bfseries Dominance} & \multicolumn{3}{>{\raggedright\arraybackslash}p{3cm}}{\bfseries Attractiveness} \\
 \cline{5-13}
  & & \multicolumn{2}{>{\raggedright\arraybackslash}p{1.5cm}|}{} & {\bfseries IS $\uparrow$} & {\bfseries PD $\uparrow$} & {\bfseries FID $\downarrow$} & {\bfseries IS $\uparrow$} & {\bfseries PD $\uparrow$} & {\bfseries FID $\downarrow$} & {\bfseries IS $\uparrow$} & {\bfseries PD $\uparrow$} & {\bfseries FID $\downarrow$} \\
 \hline
 \multirow{12}{1cm}{\rotatebox[origin=c]{90}{\specialcell{CelebAMask-HQ~\cite{CelebAMask-HQ} \\ (Real)}}} & \multirow{6}{0.7cm}{\ding{56}} & \multicolumn{2}{>{\raggedright\arraybackslash}p{1.5cm}|}{FFHQ~\cite{karras2019style}} & 0.806 & 0.262 & \phantom{0}3.28 & 0.808 & 0.284 & \phantom{0}2.94 & 0.813 & 0.275 & \phantom{0}4.24 \\
  & & \multicolumn{2}{>{\raggedright\arraybackslash}p{1.5cm}|}{Synthetic} & 0.783 & 0.263 & \phantom{0}3.22 & 0.792 & 0.281 & \phantom{0}3.05 & 0.801 & 0.268 & \phantom{0}3.81 \\
 \cline{3-13}
-  & & \multirow{4}{0.3cm}{\rotatebox[origin=c]{90}{Out-Domain}} & CFD~\cite{ma2015chicago, ma2021chicago} & 0.782 & 0.283 & 14.75 & 0.870 & 0.214 & \phantom{0}9.70 & 0.840 & 0.255 & 15.08 \\
  & & & SCUT~\cite{liang2018scut} & 0.793 & 0.265 & 14.44 & 0.831 & 0.251 & 11.84 & 0.833 & 0.258 & 14.64 \\
  & & & UAF~\cite{bainbridge2013intrinsic} & 0.782 & 0.264 & 15.05 & 0.830 & 0.245 & 10.62 & 0.787 & 0.281 & 15.62 \\
  & & & OMI~\cite{peterson2022omi} & 0.784 & 0.246 & 11.43 & 0.789 & 0.265 & 13.04 & 0.814 & 0.245 & 11.65 \\
  \cline{2-13}
  & \multirow{6}{0.7cm}{\ding{52}} & \multicolumn{2}{>{\raggedright\arraybackslash}p{1.5cm}|}{FFHQ~\cite{karras2019style}} & 0.936 & 0.159 & \phantom{0}1.42 & 0.883 & 0.223 & \phantom{0}2.36 & 0.908 & 0.199 & \phantom{0}2.59 \\
  & & \multicolumn{2}{>{\raggedright\arraybackslash}p{1.5cm}|}{Synthetic} & 0.928 & 0.158 & \phantom{0}1.29 & 0.872 & 0.220 & \phantom{0}2.21 & 0.901 & 0.193 & \phantom{0}2.28 \\
 \cline{3-13}
  & & \multirow{4}{0.3cm}{\rotatebox[origin=c]{90}{Out-Domain}} & CFD~\cite{ma2015chicago, ma2021chicago} & 0.927 & 0.154 & \phantom{0}6.90 & 0.906 & 0.186 & \phantom{0}8.62 & 0.900 & 0.211 & 11.39 \\
  & & & SCUT~\cite{liang2018scut} & 0.941 & 0.150 & \phantom{0}7.10 & 0.885 & 0.206 & 11.38 & 0.917 & 0.192 & 10.56 \\
  & & & UAF~\cite{bainbridge2013intrinsic} & 0.944 & 0.142 & \phantom{0}5.88 & 0.900 & 0.198 & \phantom{0}9.47 & 0.904 & 0.193 & \phantom{0}9.46  \\
  & & & OMI~\cite{peterson2022omi} & 0.923 & 0.155 & \phantom{0}5.92 & 0.865 & 0.215 & \phantom{0}9.21 & 0.897 & 0.185 & \phantom{0}8.09 \\
 \hline
\end{tabular}
\end{sc}
\end{small}
\end{center}
\vspace{-0.6cm}
\end{table*}

\subsection{Image Inversion of Transformed Latent}
\label{section:image_inversion_module}

The final phase allows reconstruction of the edited image latent as a transformed face. After providing image latent $w_{512}$ as input to the mapping module, it is also given as input to the generator (G) to reconstruct the image $x'$. Hence, $x' = G(w_{512}, \theta)$ (Fig.~\ref{fig:face_transform_pipeline}). We then obtain the appearance code via fusion, $h_{1024x8x8}$ (Section~\ref{section:latent_score_extraction}). To embed this information in the GAN feature space, a series of hypernetwork transformation steps~\cite{dinh2022hyperinverter} are applied to $h$ to predict residual weights --- these steps predict image-level weight updates for the generator, improving inversion quality. Finally, the transformed vector $w'$ is passed through the updated generator to reconstruct the transformed image.

\section{Experimental Details}
\label{section:experimental_details}

Our goal is to transform an input face according to a user-defined attribute score change. The pipeline is set up in three phases, to generate a transformation spectrum capable of allowing minor/major edits to decrease/increase the perception of a selected attribute. We determine that fixing the inversion and attribute prediction branches with pre-trained models helps to learn this mapping efficiently (Section~\ref{section:latent_score_extraction}). We set up FFHQ~\cite{karras2019style}, SG2-ADA~\cite{karras2020training} generated images, and CelebAMask-HQ~\cite{CelebAMask-HQ} as training data.

\begin{figure}[htbp]
\vspace{-0.1cm}
\centering
    \includegraphics[width=8.5cm]{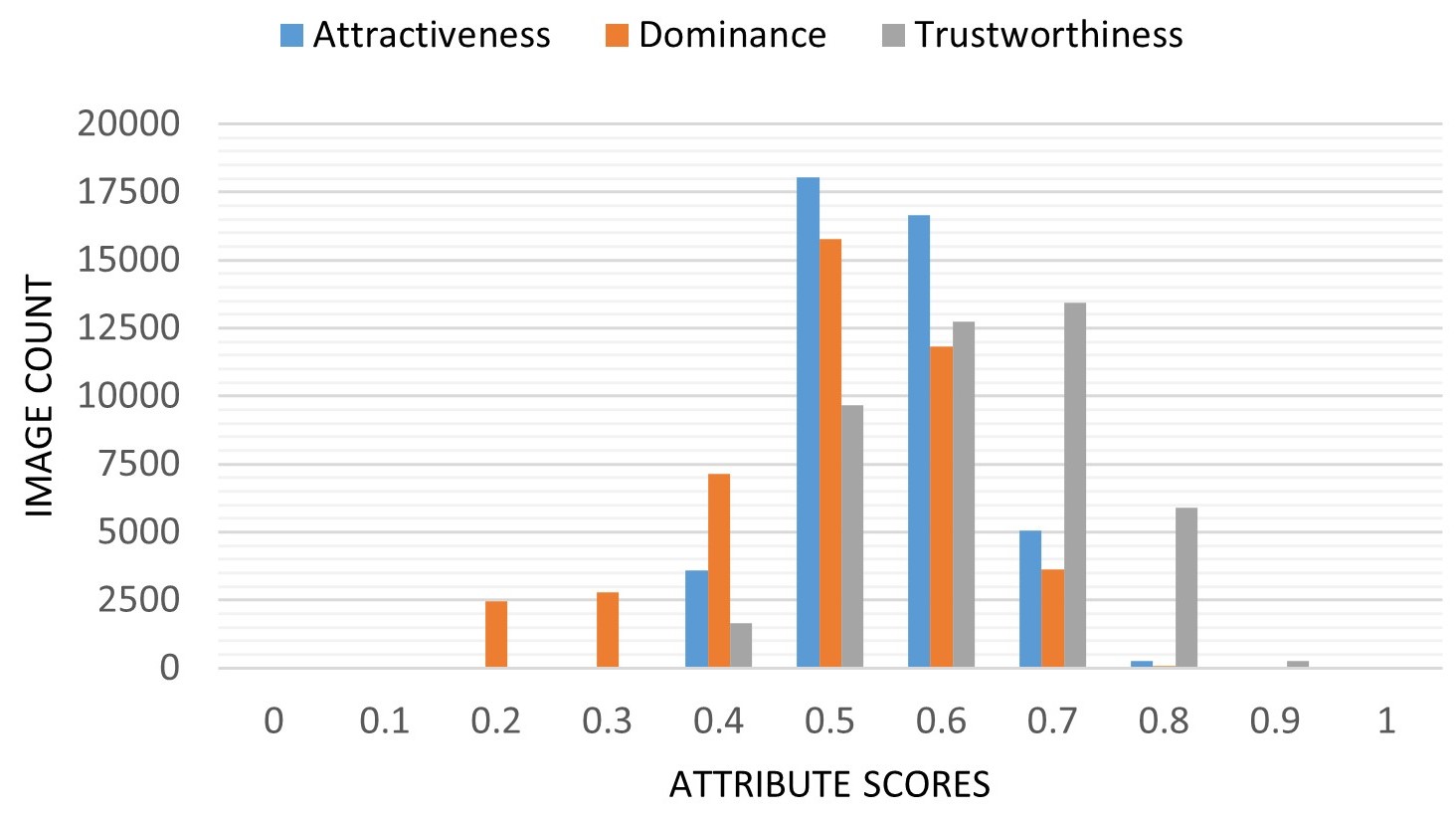}
    \vspace{-0.3cm}
    \caption{Predicted score distribution for evaluation image sets.}
\label{fig:orig_image_preds_eval_sets}

\end{figure}

\noindent\textbf{User-defined Transformation Spectrum - } We perform transformations within $[-0.4, 0.4]$ for user-defined changes, which essentially transforms images to elicit scores within $[0,1]$ --- most images in considered datasets elicit scores within $[0.4, 0.6]$, as represented in Figure~\ref{fig:orig_image_preds_eval_sets}. Some image-score pairs can provide the pipeline with a transformation target beyond $[0, 1]$, generating undesired results (Section~\ref{section:limitations_biases}). This \textit{limits our quantitative evaluations within the transformation range of $[-0.2, 0.2]$} (Tables~\ref{table:final_results},~\ref{table:final_results_id_loss},~\ref{table:id_loss_test}).

\subsection{Evaluation Metrics}
\label{section:eval_metrics}

We perform evaluations based on the following metrics.

\noindent\textit{Identity Similarity (IS):} Evaluates the \textit{identity-preserving ability}, following prior GAN-based approaches~\cite{he2022gcfsr, shoshan2021gan}. We calculate cosine similarity between ArcFace~\cite{deng2019arcface} embeddings of the original and edited images, denoted by $o$ and $e$ respectively (Eqn.~\ref{eqn:id_score_metric}), to measure identity-invariant aspects of face images. IS=1 shows perfect identity retention.

\small
\begin{equation}
IS={\frac{1}{N}} {\sum_{j=1}^{N} \frac{ \sum_{i=1}^{n}{{\bf o}_{ij}{\bf e}_{ij}} }{ \sqrt{\sum_{i=1}^{n}{({\bf o}_{ij})^2}} \sqrt{\sum_{i=1}^{n}{({\bf e}_{ij})^2}} }}
\label{eqn:id_score_metric}
\end{equation}
\normalsize
\noindent\textit{Perception Distance (PD):} Evaluates perceptual difference between original and edited images. We use modified Learned Perceptual Image Patch Similarity (LPIPS) score with VGG-16 backbone~\cite{zhang2018perceptual} (Eqn.~\ref{eqn:ps_score_metric}). PD measures the amount of transformation to input image, with higher scores implying more human-like perceptually visible edits.

\small
\begin{equation}
PD={\frac{1}{N}} {\sum_{i=1}^{N} \sum_{l} \frac{1}{H_{li}W_{li}} \sum_{h, w} || w_{li} \odot (\hat{y}_{hw}^{li} - \hat{y}_{0hw}^{li}) ||_2^2}
\label{eqn:ps_score_metric}
\end{equation}
\normalsize
\noindent\textit{Fréchet Inception Distance (FID)\footnote{https://github.com/mseitzer/pytorch-fid}:} Evaluates image quality, summarizing the distance between distributions of feature vectors of original and edited images extracted from global spatial pooling layer of Inception V3~\cite{szegedy2016rethinking} (Eqn.~\ref{eqn:fid_score_metric}).

\small
\begin{equation}
FID={\frac{1}{N}} {\sum_{i=1}^{N} \sqrt(||\mu_{1i} - \mu_{2i}||^2 + Tr(\Sigma_{1i} + \Sigma_{2i} - 2\sqrt(\Sigma_{1i}\Sigma_{2i})))}
\label{eqn:fid_score_metric}
\end{equation}
\normalsize
\noindent\textbf{Interpreting Results:} The trade-off between IS and PD scores is expected --- with perceptually visible edits (higher PD), identity of the subject probably changes (lower IS). Hence, the combined interpretation of IS ($\uparrow$) and PD ($\uparrow$) gives us the \textit{overall editing performance}. Our experiments conduct image transformations over four score changes $\{-0.2, -0.1, +0.1, +0.2\}$, corresponding to results presented in the evaluation tables. We calculate scores between each adjacent change: -0.2 with -0.1, -0.1 with original images, and so on. This allows evaluation for changes in identity, perception and quality across the transformation spectrum. Results in Tables~\ref{table:final_results_id_loss} and~\ref{table:id_loss_test} show scores averaged over these calculations.

\noindent\textbf{Visualizing Transformations:} We identify combinations of objective edits performed by the pipeline for the proposed subjective-style edits. Specific changes are identified by inverting $w_{diff}$ vectors produced from the difference between original and transformed vectors. The AF (Added Features) and RF (Removed Features) are the $w_{diff}$ vectors defined in Eqns.~\ref{eqn:AF} \&~\ref{eqn:RF} over a complete set of transformations ---

\small
\begin{equation}
AF_{ij} = {(w_i - w_j)}
\label{eqn:AF}
\end{equation}
\vspace{-0.5cm}
\begin{equation}
RF_{ji} = {(w_j - w_i)}
\label{eqn:RF}
\end{equation}
\normalsize
where $i$ corresponds to score range $[-0.3, 0.4]$ and $j$ corresponds to $[-0.4, 0.3]$. After calculating the difference vectors, the pipeline's inversion process is similarly applied to get final visualizations of the transitioning features. Please refer Appendix~\ref{section:visualizing_transformations} for visualizations and discussion.

\section{Discussion}
\label{section:overall_results}

Table~\ref{table:final_results} shows results for the proposed framework using \textit{real} and \textit{synthetic} data for training, \textit{without ID Loss}. We can observe that identities of the original faces are still largely retained (IS $>0.75$) while providing visually perceivable transformations (PD $>0.2$). The FFHQ~\cite{karras2019style} and StyleGAN2-ADA~\cite{karras2020training} generated images are in-domain, as these were either used to train the generator and/or the mapping module. The results on these datasets are predictably better. The out-of-domain results are comparatively closer to their in-domain counterparts, with a performance concern being the quality of the generated faces ($\uparrow$ FID scores). 

\begin{figure*}[htbp]
    \centering
    \includegraphics[width=18cm]{images/qualitative_performance_MAIN.jpg}
    \vspace{-0.3cm}
    \caption{Qualitative results for training \textit{without ID loss} (\textit{left}). Training data image feature diversity affecting the transformations. We observe apparent differences in edits when comparing both sides, like \textit{decrease in age} during the reduction in dominance perception or increase in trustworthiness perception, or even subtle changes like difference in hairstyles or the subject's eye color. Qualitative results \textit{with ID loss} are shown on the \textit{right}. }
\label{fig:ffhq_vs_celebamask_performance}
\vspace{-0.5cm}
\end{figure*}

The images in Figure~\ref{fig:ffhq_vs_celebamask_performance} show semantically continuous transformations on either side of the original (FFHQ Training). We observe that for edits performed with $\pm$0.2 level of changes, the faces seem to also undergo a reduction in perceived age. The scores refer to the level of desired subjective attribute change, provided during input. This was evident while targeting a reduction in dominance perception, or an increase in the trustworthiness perception. We believe there to be a correlation with the training image pool of features with the type and level of transformations we get from the pipeline. As the currently observed transformations were in correlation with decrease in age, we speculate its cause being the presence of images of children in the training set. To validate this assumption, we train models using \textit{real, adult} faces --- the CelebAMask-HQ~\cite{CelebAMask-HQ} dataset. We notice reduction in age-related manipulation of the subject with respect to a certain attribute (Figure~\ref{fig:ffhq_vs_celebamask_performance}). Although this test does not guarantee a trend with respect to the type of transformations expected from certain training setups, due to the subjectiveness of image-attribute score pairs, it does reflect on the scope of the pipeline. Training of the pipeline over any subjective attribute data is limited by the distribution of the image feature space of training data. Due to these observations, further experiments on \textit{identity loss} (Table~\ref{table:final_results_id_loss}) are performed with CelebAMask-HQ~\cite{CelebAMask-HQ}. Qualitative results are shown in Figure~\ref{fig:ffhq_vs_celebamask_performance}. Additional results are provided in Appendix~\ref{section:additional_qualitative_results}.

\subsection{Edit Consistency with Human Perception}
\label{section:edit_consistency_with_human_perception}

\begin{table}[htbp]
\vspace{-0.3cm}
\begin{center}
\begin{small}
\caption{Consistency of our face transformations with human perception. Refer Section~\ref{section:edit_consistency_with_human_perception} for details on metrics. }
\vspace{-0.15cm}
\label{table:human_study}
\begin{sc}
\renewcommand{\arraystretch}{1.1}
\begin{tabular}{ >{\raggedright\arraybackslash}p{0.4cm}| >{\raggedright\arraybackslash}p{1.25cm}| >{\raggedright\arraybackslash}p{1cm} | >{\raggedright\arraybackslash}p{0.88cm} |
>{\raggedright\arraybackslash}p{0.78cm} |>{\raggedright\arraybackslash}p{1.35cm} }
 \hline
 {\bfseries Set} & {\bfseries Attr.} & {\bfseries CWHP} & {\bfseries NVPC} & {\bfseries IWHP} & {\bfseries Disagree} \\
 \hline
 \multirow{3}{0.3cm}{\rotatebox[origin=c]{90}{\specialcell{FFHQ}}} & Trust. & 0.52 & 0.10 & 0.24 & 0.14 \\
 & Dom. & 0.67 & 0.10 & 0.14 & 0.09 \\
 & Attract. & 0.65 & 0.08 & 0.18 & 0.09 \\
 \hline
 \multirow{3}{0.3cm}{\rotatebox[origin=c]{90}{\specialcell{Synth.}}} & Trust. & 0.48 & 0.16 & 0.11 & 0.25 \\
 & Dom. & 0.58 & 0.06 & 0.25 & 0.11 \\
 & Attract. & 0.66 & 0.03 & 0.24 & 0.07 \\
 \hline
\end{tabular}
\end{sc}
\end{small}
\end{center}
\vspace{-0.4cm}
\end{table}

We conduct a human study (Appendix~\ref{section:consistency_human_perception}) with 54 participants to verify consistency of our outputs with human perception of $\uparrow$/$\downarrow$ first impression attributes (Table~\ref{table:human_study}), and evaluate based on 4 metrics. The study was IRB approved and informed consent was obtained from each participant. An image is considered Consistent with Human Perception (CWHP) if $\geq$2 (majority) participants respond similar to intended change --- if in the image pair shown, image 2 is more trustworthy, and majority felt the same. Inconsistency (IWHP) is the opposite to CWHP. If majority participants responded with \textit{can't tell the difference}, then there is No Visible Perceptual Change (NVPC) in the transformed image. If responses did not fit any of the three categories, we consider disagreement. We observe that most of our edits are visible to the human participants, with a small fraction (NVPC $\leq$10\%) failing to provide meaningful perceptual changes. Edits for Dominance and Attractiveness attributes show higher agreement, with CWHP $>$55\% for our original-edit image pairs. We can observe from Table~\ref{table:final_results_id_loss} that both IS and PD for these attributes are high, which shows that edits are semantically significant with respect to human judgment. This might mean \textit{human biases for these attributes are stronger} as well, further discussed in Section~\ref{section:limitations_biases}. 

\subsection{Identity Preservation Ability}
\label{section:identity_preservation}

\vspace{-0.3cm}
\begin{table}[htbp]
\begin{center}
\begin{small}
\caption{Demonstrating the improvement in identity preservation using \textit{ID Loss}. PED: Principal Editing Direction. Attribute: Dominance. Training iterations: $\sim$15190.}
\vspace{-0.15cm}
\label{table:id_loss_test}
\begin{sc}
\renewcommand{\arraystretch}{1.1}
\begin{tabular}{ >{\raggedright\arraybackslash}p{2cm}| >{\raggedright\arraybackslash}p{2.25cm}| >{\raggedright\arraybackslash}p{1cm} }
 \hline
 {\bfseries Train Set} & {\bfseries Freeze PED} & {\bfseries IS $\uparrow$} \\
 \hline
 \multirow{4}{0.3cm}{FFHQ} & \textbf{4} & \textbf{0.932} \\
 & 12 & 0.909 \\
 & 27 & 0.908 \\
 & No ID Loss & 0.894 \\
 \hline
\end{tabular}
\end{sc}
\end{small}
\end{center}
\vspace{-0.5cm}
\end{table}

We demonstrate improvement in identity preservation using \textit{Identity Loss} (Section~\ref{section:mapping_module}) in Table~\ref{table:id_loss_test}. Freezing first 4 PEDs of the target latent with identity loss training provides the highest performance improvement. Comparing with non-ID Loss variants in Table~\ref{table:final_results_id_loss}, we observe \textit{significant improvements in identity retention ($\uparrow$ IS scores), while largely maintaining perception changes (PD $>0.14$)}.

\begin{table}[htbp]
\begin{center}
\begin{small}
\caption{Comparison with OMI~\cite{peterson2022omi} on released validation images. \textit{Real} images are from CFD while Synthetic (Syn.) images are from OMI's artificial released subset.}
\vspace{-0.15cm}
\label{table:omi_comparison}
\begin{sc}
\renewcommand{\arraystretch}{1.2}
\begin{tabular}{ >{\raggedright\arraybackslash}p{0.7cm}| >{\raggedright\arraybackslash}p{1.25cm}| >{\raggedright\arraybackslash}p{0.88cm} | >{\raggedright\arraybackslash}p{0.88cm} | >{\raggedright\arraybackslash}p{0.88cm} | >{\raggedright\arraybackslash}p{0.88cm} }
 \hline
 {\bfseries Edits} & {\bfseries Attr.} & \multicolumn{2}{c|}{\bfseries IS} & \multicolumn{2}{c}{\bfseries PD} \\
 \hline
 & & Real & Syn. & Real & Syn. \\
 \hline
 \multirow{3}{0.3cm}{\rotatebox[origin=c]{90}{\specialcell{OMI}}} & Trust. & 0.962 & 0.908 & 0.095 & 0.193 \\
 & Dom. & 0.957 & 0.880 & 0.110 & 0.231 \\
 & Attract. & 0.962 & 0.880 & 0.119 & 0.255 \\
 \hline
 \multirow{3}{0.3cm}{\rotatebox[origin=c]{90}{\specialcell{Ours}}} & Trust. & 0.935 & 0.922 & 0.142 & 0.161 \\
 & Dom. & 0.894 & 0.868 & 0.204 & 0.214 \\
 & Attract. & 0.896 & 0.891 & 0.211 & 0.202 \\
 \hline
\end{tabular}
\end{sc}
\end{small}
\end{center}
\vspace{-0.75cm}
\end{table}

To verify identity retention across $[-0.2, 0.2]$, we use images generated from the \textit{ID Loss} training setup to perform a second human study (Appendix~\ref{section:human_study_id_preservation}) --- participants agree that our transformations retain identity for 73\%, 60\%, and 86\% of the face pairs for \textit{attractiveness}, \textit{dominance}, and \textit{trustworthiness}, respectively. We also compare results to Peterson \etal~\cite{peterson2022omi}. They use vector arithmetic over learned weights from linear regression models to edit latent vectors. While their transformations are smooth, they are unable to properly disentangle the feature space from the image identities. Table~\ref{table:omi_comparison} shows comparison of both methods on the validation sets released by~\cite{peterson2022omi} --- our transformations are consistent over both real and synthetic faces, and competitively more perceivable (higher PD). Note that while their model is trained with \textit{synthetic} data in Table~\ref{table:omi_comparison}, our results provide higher IS scores for \textit{trustworthiness} and \textit{attractiveness} (Qualitative results in Appendix~\ref{section:omi_comparison}).

\section{Synthetic Data to Improve First Impression Prediction: An Application}

We now demonstrate that synthetic face augmentation, using our proposed method, improves the performance and out-of-domain generalizability of facial first impression prediction models. Our goal is to train human-like first impression prediction models on attribute-specific features.

\begin{table*}[htbp]
\begin{center}
\begin{small}
\caption{Comparing performances of attribute prediction modules \textit{before} and \textit{after} augmenting the training data with generated images over a transformation spectrum of $[-0.2, 0.2]$ for an attribute. ResNet-50~\cite{he2016deep} architecture is used. Evaluation sets are shown as columns. $R^2$ ($\uparrow$) (coefficient of determination) scores are reported.}
\vspace{-0.15cm}
\label{table:final_scores_resnet50}
\begin{sc}
\renewcommand{\arraystretch}{1.3}
\begin{tabular}{ >{\raggedright\arraybackslash}p{1cm} |
>{\raggedright\arraybackslash}p{3cm} |>{\raggedright\arraybackslash}p{0.85cm} >
{\raggedright\arraybackslash}p{0.85cm} >{\raggedright\arraybackslash}p{0.85cm} >{\raggedright\arraybackslash}p{0.85cm} |>{\raggedright\arraybackslash}p{0.85cm} >{\raggedright\arraybackslash}p{0.85cm} |>{\raggedright\arraybackslash}p{0.85cm} >{\raggedright\arraybackslash}p{0.85cm} >{\raggedright\arraybackslash}p{0.85cm} }
 \hline
 \multirow{2}{1cm}{\bfseries Train} & {\multirow{2}{3cm}{\bfseries Data Augment}} & \multicolumn{4}{>{\raggedright\arraybackslash}p{3.4cm}|}{\bfseries Attractiveness} & \multicolumn{2}{>{\raggedright\arraybackslash}p{1.7cm}}{\bfseries Dominance} & \multicolumn{3}{|>{\raggedright\arraybackslash}p{2.55cm}}{\bfseries Trustworthiness} \\
 \cline{3-11}
  & & {\bfseries CFD} & {\bfseries OMI} & {\bfseries UAF} & {\bfseries SCUT} & {\bfseries CFD} & {\bfseries OMI} & {\bfseries CFD} & {\bfseries OMI} & {\bfseries UAF} \\
 \hline
 \multirow{3}{1cm}{\rotatebox[origin=c]{90}{\specialcell{CFD}}} & {Original (Real)} & 0.399 & 0.324 & 0.018 & 0.001 & 0.314 & 0.163 & \textbf{0.275} & 0.038 & 0.017 \\
  & {Original + Syn\_\scriptsize{ID}} & 0.413 & 0.473 & 0.225 & 0.111 & \textbf{0.458} & 0.320 & 0.268 & 0.451 & 0.137 \\
  & Original + Syn\_\scriptsize{OD} & \textbf{0.454} & \textbf{0.580} & \textbf{0.395} & \textbf{0.322} & 0.436 & \textbf{0.850} & 0.248 & \textbf{0.830} & \textbf{0.430} \\
  \hline
  \multirow{3}{1cm}{\rotatebox[origin=c]{90}{\specialcell{OMI}}} & {Original (Real)} & 0.317 & 0.595 & 0.473 & 0.196 & 0.065 & 0.798 & 0.127 & 0.699 & 0.383 \\
  & {Original + Syn\_\scriptsize{ID}} & \textbf{0.353} & \textbf{0.647} & \textbf{0.473} & 0.251 & 0.182 & 0.831 & 0.139 & 0.810 & \textbf{0.420} \\
  & Original + Syn\_\scriptsize{OD} & 0.281 & 0.609 & 0.321 & \textbf{0.331} & \textbf{0.341} & \textbf{0.868} & \textbf{0.141} & \textbf{0.852} & 0.412 \\
  \hline
  \multirow{3}{1cm}{\rotatebox[origin=c]{90}{\specialcell{UAF}}} & {Original (Real)} & 0.436 & 0.487 & 0.574 & 0.366 & - & - & \textbf{0.262} & 0.271 & 0.612 \\
  & {Original + Syn\_\scriptsize{ID}} & \textbf{0.446} & 0.531 & \textbf{0.588} & \textbf{0.381} & - & - & 0.211 & 0.545 & \textbf{0.623} \\
  & Original + Syn\_\scriptsize{OD} & 0.350 & \textbf{0.599} & 0.545 & 0.301 & - & - & 0.200 & \textbf{0.799} & 0.592 \\
  \hline
  \multirow{3}{1cm}{\rotatebox[origin=c]{90}{\specialcell{SCUT}}} & {Original (Real)} & 0.272 & 0.214 & 0.310 & \textbf{0.781} & - & - & - & - & - \\
  & {Original + Syn\_\scriptsize{ID}} & \textbf{0.334} & 0.392 & \textbf{0.324} & 0.761 & - & - & - & - & - \\
  & Original + Syn\_\scriptsize{OD} & 0.299 & \textbf{0.582} & 0.277 & 0.728 & - & - & - & - & - \\
 \hline
\end{tabular}
\end{sc}
\end{small}
\end{center}
\vspace{-0.6cm}
\end{table*}

\subsection{Data Setup}

\begin{table}[htbp]
\begin{center}
\begin{small}
\caption{Syn. gen. images / annotations using our proposed pipeline. \textit{E\scriptsize{XP}.} indicates the training setup for which the synthetic data is utilized. Gen. images distributed equally over score changes $[-0.2,0.2]$. Keys --- ID: In-Domain, OD: Out-of-Domain.}
\vspace{-0.3cm}
\label{table:generated_datasets}
\begin{sc}
\renewcommand{\arraystretch}{1.3}
\begin{tabular}{ >{\raggedright\arraybackslash}p{1cm} |>{\raggedright\arraybackslash}p{3cm} |
>{\raggedright\arraybackslash}p{3cm}}
 \hline
{\bfseries Exp.} & {\bfseries Dataset} & {\bfseries Total Gen. Images} \\
 \hline
\multirow{4}{0.5cm}{Syn\_\scriptsize{ID}} & CFD & 1984 \\
 & UAF & 2384 \\
 & SCUT & 3100 \\
 & OMI & 2392 \\
 \hline
 \multirow{2}{0.5cm}{Syn\_\scriptsize{OD}} & FFHQ~\cite{karras2019style} & 17492 \\
 & SG2-ADA Gen.~\cite{karras2020training} & 17492 \\
 \hline
\end{tabular}
\end{sc}
\end{small}
\end{center}
\vspace{-0.9cm}
\end{table}
 
We select four popular datasets for synthetic data augmentation --- Chicago Face Dataset~\cite{ma2015chicago, ma2021chicago}, One Million Impressions Dataset~\cite{peterson2022omi}, US Adult Faces~\cite{bainbridge2013intrinsic}, and SCUT-FBP5500~\cite{liang2018scut}\footnote{Details of the original datasets are described in Appendix~\ref{section:improve_fi_prediction_supp}.}. We split each original dataset by 60-40 (\%), for training and evaluation respectively. Each evaluation set is used to test all trained models for the attributes, providing out-of-domain performance. Using the training splits, we generate synthetic images for attribute score changes within $[-0.2,0.2]$ using our proposed method (CelebAMask-HQ~\cite{CelebAMask-HQ} trained model). This introduces variability in face image stimuli for a particular identity. We term this augmentation as \textit{in-domain}, as variations of identities within the same dataset are generated. Details of augmentations are shown in Table~\ref{table:generated_datasets}.

We also include an \textit{out-of-domain} variation, investigating whether adding synthetic face variations of new identities within the training sets improves performance on the fixed validation sets. We sample 4373 real and synthetic faces each, from FFHQ~\cite{karras2019style} and synthetic images generated by StyleGAN2-ADA~\cite{karras2020training}, respectively. We then augment these sets using synthetic variations generated using our method. The now available 17492 samples from both these sets are appended to all four training datasets. For both the \textit{in-domain} and \textit{out-of-domain} augmentations, attribute annotations are not available --- we generate them using our score prediction models (Appendix~\ref{section:subjective_attribute_prediction}). 

\subsection{Model Training}

Inspired by~\cite{mccurrie2018convolutional}, we train the ResNet-50~\cite{he2016deep} architecture for the subjective first impression prediction task. We use a constant set of training hyperparameters for all experiments; these include training over a batch size of 16 with a learning rate of 6x$10^{-5}$ for 250 epochs. All input images are resized to 224x224. To show the generalizability of using such a data augmentation method in improving model performance for subjective tasks, we conduct similar experiments on the SWIN-Tiny variant~\cite{liu2021swin}. This architecture choice is based on the general improvement of transformer-based models over state-of-the-art CNN-based methods for traditional objective visual tasks~\cite{dosovitskiy2020image}, and the ability of SWIN~\cite{liu2021swin} models to capture human-relevant features for impression prediction. Additionally, the \textit{Tiny} variant's number of parameters (29M) are comparable to the ResNet-50 architecture (23.9M).
 
\begin{table*}[htbp]
\begin{center}
\begin{small}
\caption{Comparing performances of attribute prediction modules with SWIN-Tiny~\cite{liu2021swin} architecture, setup similar to Table~\ref{table:final_scores_resnet50}. 
} 
\vspace{-0.3cm}
\label{table:final_scores_vit}
\begin{sc}
\renewcommand{\arraystretch}{1.3}
\begin{tabular}{ >{\raggedright\arraybackslash}p{1cm} |
>{\raggedright\arraybackslash}p{3cm} |>{\raggedright\arraybackslash}p{0.85cm} |>
{\raggedright\arraybackslash}p{0.85cm} |>{\raggedright\arraybackslash}p{0.85cm} |>{\raggedright\arraybackslash}p{0.85cm} |>{\raggedright\arraybackslash}p{0.85cm} |>{\raggedright\arraybackslash}p{0.85cm} |>{\raggedright\arraybackslash}p{0.85cm} |>{\raggedright\arraybackslash}p{0.85cm} |>{\raggedright\arraybackslash}p{0.85cm} }
 \hline
 \multirow{2}{1cm}{\bfseries Train} & {\multirow{2}{3cm}{\bfseries Data Augment}} & \multicolumn{4}{>{\raggedright\arraybackslash}p{3.4cm}|}{\bfseries Attractiveness} & \multicolumn{2}{>{\raggedright\arraybackslash}p{1.7cm}}{\bfseries Dominance} & \multicolumn{3}{|>{\raggedright\arraybackslash}p{2.55cm}}{\bfseries Trustworthiness} \\
 \cline{3-11}
  & & {\bfseries CFD} & {\bfseries OMI} & {\bfseries UAF} & {\bfseries SCUT} & {\bfseries CFD} & {\bfseries OMI} & {\bfseries CFD} & {\bfseries OMI} & {\bfseries UAF} \\
 \hline
 \multirow{2}{1cm}{\rotatebox[origin=c]{90}{\specialcell{CFD}}} & {Original (Real)} & 0.364 & 0.322 & 0.142 & 0.046 & 0.324 & 0.118 & 0.210 & 0.000 & 0.005 \\
  & {Original + Syn\_\scriptsize{ID}} & \textbf{0.386} & \textbf{0.463} & \textbf{0.317} & \textbf{0.104} & \textbf{0.405} & \textbf{0.550} & \textbf{0.340} & \textbf{0.669} & \textbf{0.461} \\
  \hline
  \multirow{2}{1cm}{\rotatebox[origin=c]{90}{\specialcell{OMI}}} & {Original (Real)} & 0.242 & 0.536 & 0.210 & 0.029 & 0.205 & 0.776 & 0.143 & 0.766 & 0.367 \\
  & {Original + Syn\_\scriptsize{ID}} & \textbf{0.385} & \textbf{0.657} & \textbf{0.414} & \textbf{0.283} & \textbf{0.289} & \textbf{0.828} & \textbf{0.192} & \textbf{0.825} & \textbf{0.435} \\
  \hline
  \multirow{2}{1cm}{\rotatebox[origin=c]{90}{\specialcell{UAF}}} & {Original (Real)} & 0.365 & 0.482 & 0.560 & 0.334 & - & - & 0.146 & 0.259 & 0.571 \\
  & {Original + Syn\_\scriptsize{ID}} & \textbf{0.371} & \textbf{0.518} & \textbf{0.580} & \textbf{0.416} & - & - & \textbf{0.223} & \textbf{0.701} & \textbf{0.597} \\
  \hline
  \multirow{2}{1cm}{\rotatebox[origin=c]{90}{\specialcell{SCUT}}} & {Original (Real)} & 0.235 & 0.160 & 0.288 & \textbf{0.747} & - & - & - & - & - \\
  & {Original + Syn\_\scriptsize{ID}} & \textbf{0.392} & \textbf{0.349} & \textbf{0.326} & 0.724 & - & - & - & - & - \\
 \hline
\end{tabular}
\end{sc}
\end{small}
\end{center}
\vspace{-0.4cm}
\end{table*}

\begin{figure*}[htbp]
    \centering
    \includegraphics[width=15cm]{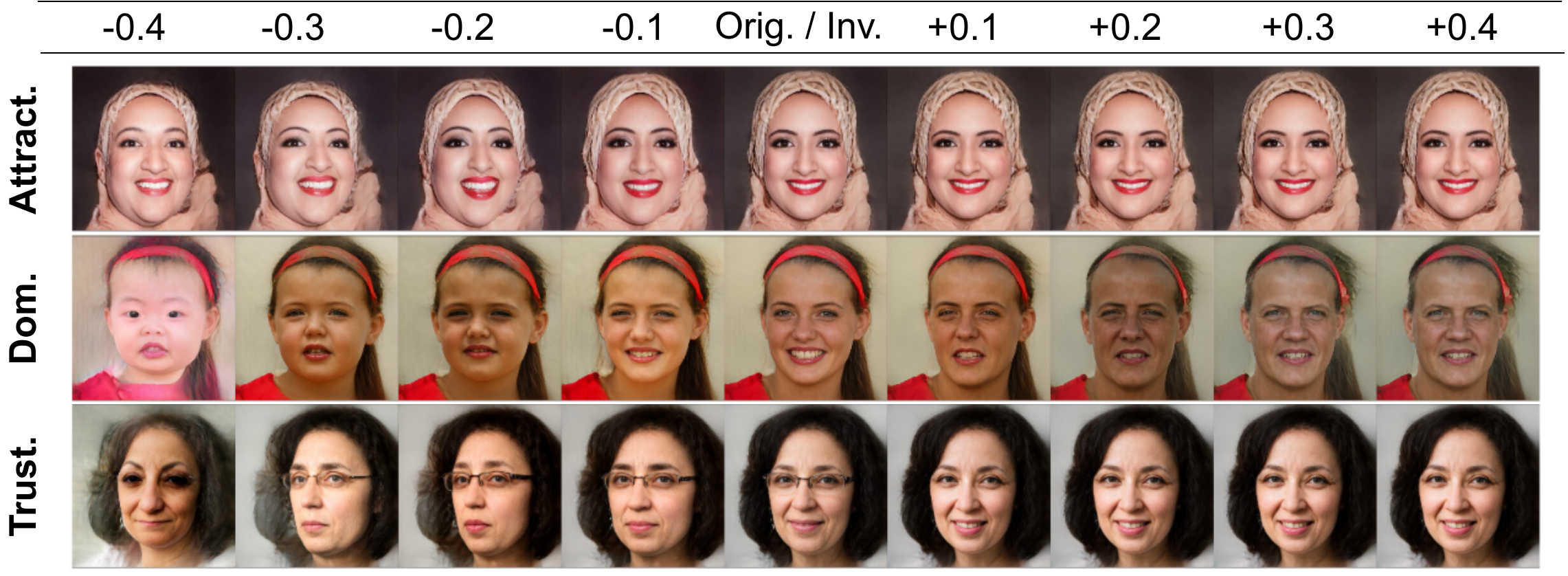}
    \vspace{-0.3cm}
    \caption{Unintended reflection of training data biases. \textit{Chubby face} ($\downarrow$ Attractiveness) and \textit{darker skin tone} ($\uparrow$ Dominance) reflect human annotation biases. Biases are further discussed in Section~\ref{section:limitations_biases}.}
\label{fig:fp_performance_limitations}
\vspace{-0.2cm}
\end{figure*}

\subsection{Generalizability and Performance Improvement}

Results for the ResNet-50~\cite{he2016deep} architecture are presented in Table~\ref{table:final_scores_resnet50}. While the \textit{Original} training represents models trained on original training splits of the dataset, the \textit{Syn\_\scriptsize{ID}} (ID: In-Domain) variants use synthetic augmentation from only the originally selected dataset, and the \textit{Syn\_\scriptsize{OD}} (OD: Out-of-Domain) variants use the FFHQ~\cite{karras2019style} and SG2-ADA~\cite{karras2020training} generated image-score pairs. We observe from the scores that the synthetic data augmentation improves performance in both cases. Additionally, due to the synthetic nature of the OMI~\cite{peterson2022omi} dataset, the performance on its evaluation set for each data augmentation is significantly better due to a better domain match than real images. These results show that the model is \textit{able to learn attribute-specific features} during training from a variety of image-score pairs for similar identities, simultaneously reducing attention towards identity-specific face features. This is represented by two notable results --- the ResNet-50 trained on both the CFD trustworthiness data and the SCUT-FBP5500 attractiveness data. In both these cases, the in-domain evaluation set performance drops slightly to compensate for higher scores on out-of-domain evaluation sets. Results for the SWIN-Tiny~\cite{liu2021swin} architecture show a similar trend (Table~\ref{table:final_scores_vit}), which demonstrates that the utility of synthetic data augmentation for subjective first impression prediction is not limited to a specific architecture.

\section{Limitations \& Training Data Biases}
\label{section:limitations_biases}

Our pipeline maps possible semantic variations of a face with respect to a subjective attribute. This allows us to generate variants of any face image according to user-desired changes in attribute score. Our range of user provided changes is between $[-0.4, 0.4]$, which roughly covers the entire transformation spectrum for editing the faces. However, some edits tend to give undesired results. An example can be observed for Trustworthiness ($-0.4$) in Figure~\ref{fig:fp_performance_limitations}, with the edit largely transforming the identity and adding unwanted artifacts to the face. A related observation for a portion of the evaluative results was the inability of the pipeline to provide face transformations with \textit{identifiable edits} between $[0.2, 0.4]$ (Figure~\ref{fig:fp_performance_limitations}, Attractiveness \& Trustworthiness). Both of these can be attributed to the possible transformation target beyond $[0, 1]$ (discussed in Section~\ref{section:experimental_details}). Such targets can be bounded within $[0, 1]$, with (+/-) $0.4$ possibly transforming to the nearest lower/upper changes, respectively.

Additionally, the pipeline tends to provide biased transformations, like adding \textit{baby face} to represent $\uparrow$ trustworthiness and $\downarrow$ dominance (Figure~\ref{fig:fp_performance_limitations}, Dominance $[-0.4, -0.2]$). We attribute this to training data (FFHQ~\cite{karras2019style}) consisting of young faces being labeled with $\uparrow$ trustworthiness and $\downarrow$ dominance (Figure~\ref{fig:ffhq_vs_celebamask_performance}). We \textit{verify and reduce unintended bias by training another set of mapping networks with a less biased range of faces~\cite{CelebAMask-HQ}} 
(Table~\ref{table:final_results_id_loss}). The most recurring biases are the association of a chubby face and darker skin tone~\cite{hutchings2024racial} with lower trustworthiness and attractiveness (Figure~\ref{fig:fp_performance_limitations}, Attractiveness). We highlight the importance of such a bias \textit{negatively affecting the interpersonal interactions}, and suggest crafting appropriate guidelines before implementing the proposed solution for any decision-making tasks.

\section{Summary \& Ethical Implications}

In this work, we demonstrate that subjective traits can be effectively manipulated using generative techniques by leveraging a learned latent space designed for subjective attribute representation. Edits performed using our proposed approach align with consensus-based human judgments, resulting in visible perceptual changes while largely preserving identity. Although some information loss may occur during image inversion, the inversion process enables critical comparisons between real and synthetic image editing by projecting images into the learned latent space and reconstructing edited latent vectors back into the image space.

Additionally, we demonstrate the utility of synthetically generated face variations across an attribute's score spectrum to enhance the performance of prediction models for perceived traits such as trustworthiness, dominance, and attractiveness. Notably, while synthetic data augmentation improves performance on real-world validation sets, a small synthetic dataset alone may fall short of achieving optimal results due to discrepancies between real and synthetic data distributions~\cite{jacobsen2023machine}. This underscores the value of synthetic data augmentation primarily for pretraining~\cite{zeng2019df2net} or regularizing tasks~\cite{kowalski2020config}, as exemplified in our approach.

There is potential for images generated by this work to be used to counteract biased judgments and misunderstandings of perceived impression as real, but the nature of the problem still encodes human biases. To understand its application, one must generate a group of images and understand the discrepancies and limits of such techniques before punctuating their use. We have shown some limitations through experiments with only adult faces, examples with low quality edits and not well-preserved identity scenarios. The proposed work intends to provide tools to understand these biases better and act as inspiration for future work to capture the full subjectivity of the problem. Code, data, and trained models will be shared for future research.

{\small
\bibliographystyle{ieee_fullname}
\bibliography{main}
}

\appendix

\onecolumn
\begin{appendix}

{\bfseries\LARGE Appendix\par}

\section{Training \& Evaluation (Section \ref{section:experimental_details})}
\label{section:training_evaluation}

\subsection{Choice of Image Inversion Module}
\label{section:image_inversion}

\begin{figure}[htbp]
    \centering
    \includegraphics[width=17cm]{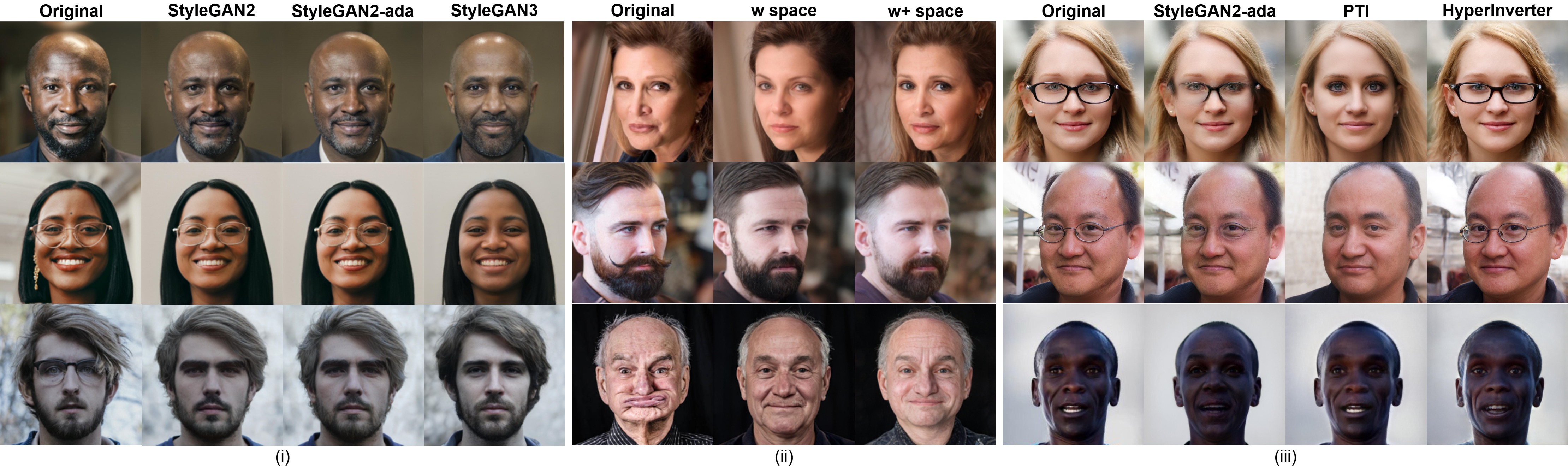}
    \caption{Inversion module selection process.}
    \label{fig:gan_arch_selection}
\end{figure}

The inversion process depends on generator architecture and type of latent space on which the input image is projected. A literature survey allowed us to choose StyleGAN \cite{karras2019style, karras2020analyzing, karras2020training, Karras2021} series of models as our selection pool for base generator architecture. These models are capable of producing high quality images, and can be included in our pipeline efficiently. We performed ablation experiments on 1000 FFHQ images and out-of-domain internet images to select the suitable method for inversion in our pipeline. The additional data was required to compare the inversion quality of the selected methods on unseen images, showing method generalizability. (i) in Fig. \ref{fig:gan_arch_selection} shows the comparison of the StyleGAN architectures in the $w$ latent space, StyleGAN2-ADA \cite{karras2020training} provides better inversions. As our selected data are random out-of-domain images, these results could differ from those published in prior papers. After selecting the StyleGAN2-ADA as our base architecture, we conduct a comparison between the highly editable $w$ space and the identity retaining $w$+ space in (ii) of Fig. \ref{fig:gan_arch_selection}. While $w$+ space results are marginally better, it gives us relatively unstable edits~\cite{abdal2021styleflow, tov2021designing}, so we continue using the $w$ space. In (iii) of Fig. \ref{fig:gan_arch_selection}, we evaluate different inversions -- PTI~\cite{roich2022pivotal} and HyperInverter~\cite{dinh2022hyperinverter}, for StyleGAN2-ADA's $w$ space. While the original StyleGAN2-ADA method uses only the $w$ vector for inversion, the PTI \cite{roich2022pivotal} and HyperInverter \cite{dinh2022hyperinverter} pipelines either finetune the generator for every new image or use additional vectors to store image information missing from the w vector, respectively. PTI and HyperInverter perform similarly on most out-of-domain images, so we conducted an additional quality test on the 1000 image FFHQ subset. We also performed a timing test to verify a better modular fit for the pipeline --- HyperInverter took 739 seconds against PTI's 194811 seconds to invert the 1000 image subset, on an NVIDIA RTX A5000 GPU. We select the HyperInverter \cite{dinh2022hyperinverter} method for its generalizable inversion performance and higher identity retention, along with better computational efficiency.

\subsection{Subjective Attribute Prediction Model}
\label{section:subjective_attribute_prediction}

We use three datasets to train the subjective attribute prediction models --- McCurrie \etal's~\cite{mccurrie2018convolutional} AFLW~\cite{koestinger11a} image subset with annotations for Dominance and Trustworthiness attributes, the SCUT-FBP5500~\cite{liang2018scut} dataset with Attractiveness data, and the One Million Impressions (OMI)~\cite{peterson2022omi} dataset with annotations for all three attributes. We pre-train models using McCurrie \etal's subset for Dominance and Trustworthiness with 6300 images, and the SCUT-FBP5500 for Attractiveness with 5500 images, and fine-tune using the OMI dataset of 1004 images --- this provides better generalizability across the datasets sampled for training, as shown in Table~\ref{table:subjective_attribute_prediction_model_training}. Some of the training datasets are described in Table~\ref{table:original_datasets}.

\begin{table}[htbp]
\begin{center}
\begin{small}
\caption{Subjective attribute models are trained using AFLW \cite{mccurrie2018convolutional, koestinger11a}, SCUT-FBP5500 \cite{liang2018scut}, and OMI \cite{peterson2022omi} datasets. We use ResNet-18 \cite{he2016deep} as base architecture, trained individually for each attribute using weighted averages of crowd ratings for each image. Models trained with data in the first row are fine-tuned with data in the second row. Models corresponding to scores in bold are used for first impression prediction of respective attributes in the pipeline. Metric: $R^2$ ($\uparrow$) (coefficient of determination).}
\label{table:subjective_attribute_prediction_model_training}
\begin{sc}
\renewcommand{\arraystretch}{1.1}
\begin{tabular}{ >{\raggedright\arraybackslash}p{2.55cm} |
>{\raggedright\arraybackslash}p{0.84cm} | >{\raggedright\arraybackslash}p{0.86cm} | >{\raggedright\arraybackslash}p{0.86cm} |
>{\raggedright\arraybackslash}p{0.86cm}}
 \hline
 {\bfseries Attribute} & {\bfseries Train data} & {\bfseries Test: AFLW} & {\bfseries Test: SCUT} & {\bfseries Test: OMI} \\
 \hline
 \multirow{2}{2.55cm}{Trustworthiness} & AFLW & 0.5234 & - & 0.3329 \\
  & \textbf{OMI} & \textbf{0.3239} & - & \textbf{0.8627} \\
 \hline
 \multirow{2}{2.55cm}{Dominance} & AFLW & 0.6012 & - & 0.1401 \\
  & \textbf{OMI} & \textbf{0.4495} & - & \textbf{0.8077} \\
 \hline
 \multirow{2}{2.55cm}{Attractiveness} & SCUT & - & 0.7302 & 0.0231 \\
  & \textbf{OMI} & - & \textbf{0.5974} & \textbf{0.5618} \\
 \hline
\end{tabular}
\end{sc}
\end{small}
\end{center}
\end{table}

\subsection{Attribute \& Image Data}
\label{section:attribute_and_image_data}

For the end-to-end face image attribute transformation pipeline, we primarily focus on training the flow mapping network inspired from~\cite{abdal2021styleflow}. The network maps image latents to attribute scores, and we train individual attribute models upto 80000 iterations with a batch size of 50 image-score pairs (learning rate = 1e-3). Our experiments used 4 stacked CNF blocks for all model training experiments. We selected model checkpoints with the lowest loss during training. Training hyperparameters are empirically selected based on initial tests.

The FFHQ~\cite{karras2019style} dataset represents \textit{real} images, which consists of 70000 images split into 50000 for training and 20000 for evaluation. Due to empirical limitations of the inversion module, and to improve the pipeline performance, we set up a filtering criteria for the training set using HyperInverter~\cite{dinh2022hyperinverter}. This criteria is designed to provide the pipeline with a diverse set of features from images that are easily invertible and have a higher face detection accuracy in the proposed end-to-end setup, discarding images that could fail in any of the presented sub-modules (Main Paper Section 3). The first criteria allows images that retain basic face characteristics during inversion --- if the inverted faces are detected with a probability $>0.999995$ using the MTCNN face detector~\cite{zhang2016joint}. For the second criteria, we use cosine similarity between the ArcFace~\cite{deng2019arcface} embeddings of the original images and their inversions before edits to obtain images whose identity is reasonably restored after inversion (identity similarity $>0.85$). The thresholds for both the criteria are empirically derived. After filtering, our final dataset consists of 10883 images for training. Although the filtering process allowed us to discard a larger number of stimuli from the training set, evaluations are conducted on the held-out set of 19921\footnote{79 images out of 20000 were dropped due to low face detection accuracy.} images.

We similarly set up two additional datasets --- synthetically generated StyleGAN2-ADA~\cite{karras2020training} images for the same training (10883) and evaluation (19921) image count, and a second real image training set filtered from the CelebAMask-HQ~\cite{CelebAMask-HQ} dataset (10883 images). To evaluate generalizability of the proposed pipeline, we utilize additional subsets of images from four datasets --- 827 images from the Chicago Face Dataset~\cite{ma2015chicago}, and 1000 images each from SCUT-FBP5500~\cite{liang2018scut}, US Adult Faces Database~\cite{bainbridge2013intrinsic}, and One Million Impressions Dataset~\cite{peterson2022omi}.

\newpage
\subsection{Score Distributions of the Subjective Attributes in Training Data}

\begin{figure}[htbp]
    \centering
    \vspace{0.25cm}
    \includegraphics[width=17cm]{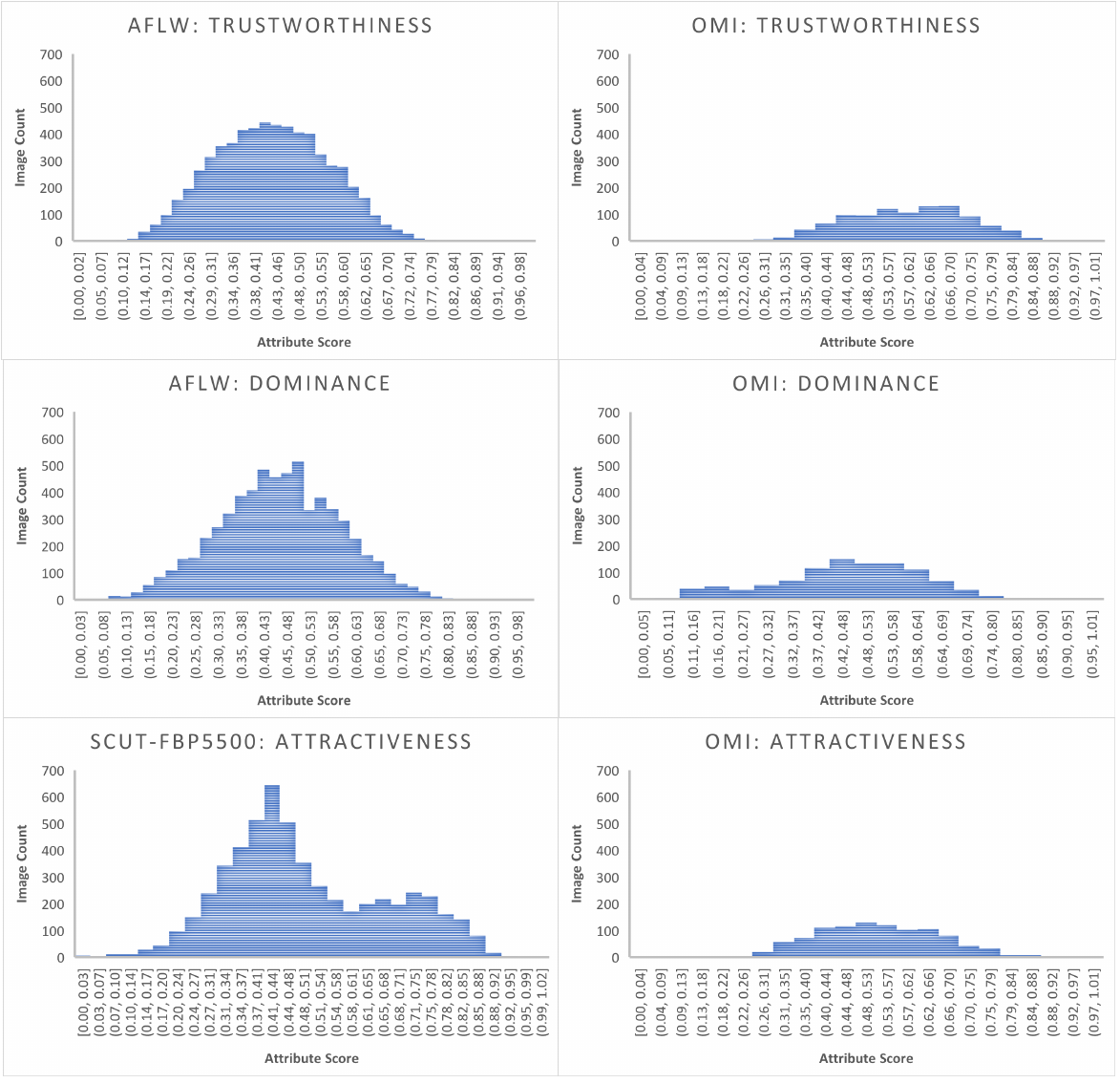}
    \caption{Subjective Data Score Distribution. AFLW and SCUT datasets contain real images and OMI contains synthetic images.}
\label{fig:subjective_data_score_distributions}
    \vspace{-0.5cm}
\end{figure}

\newpage
\subsection{Visualizing Transformations}
\label{section:visualizing_transformations}
\begin{figure*}[htbp]
    \centering
    \includegraphics[width=15cm]{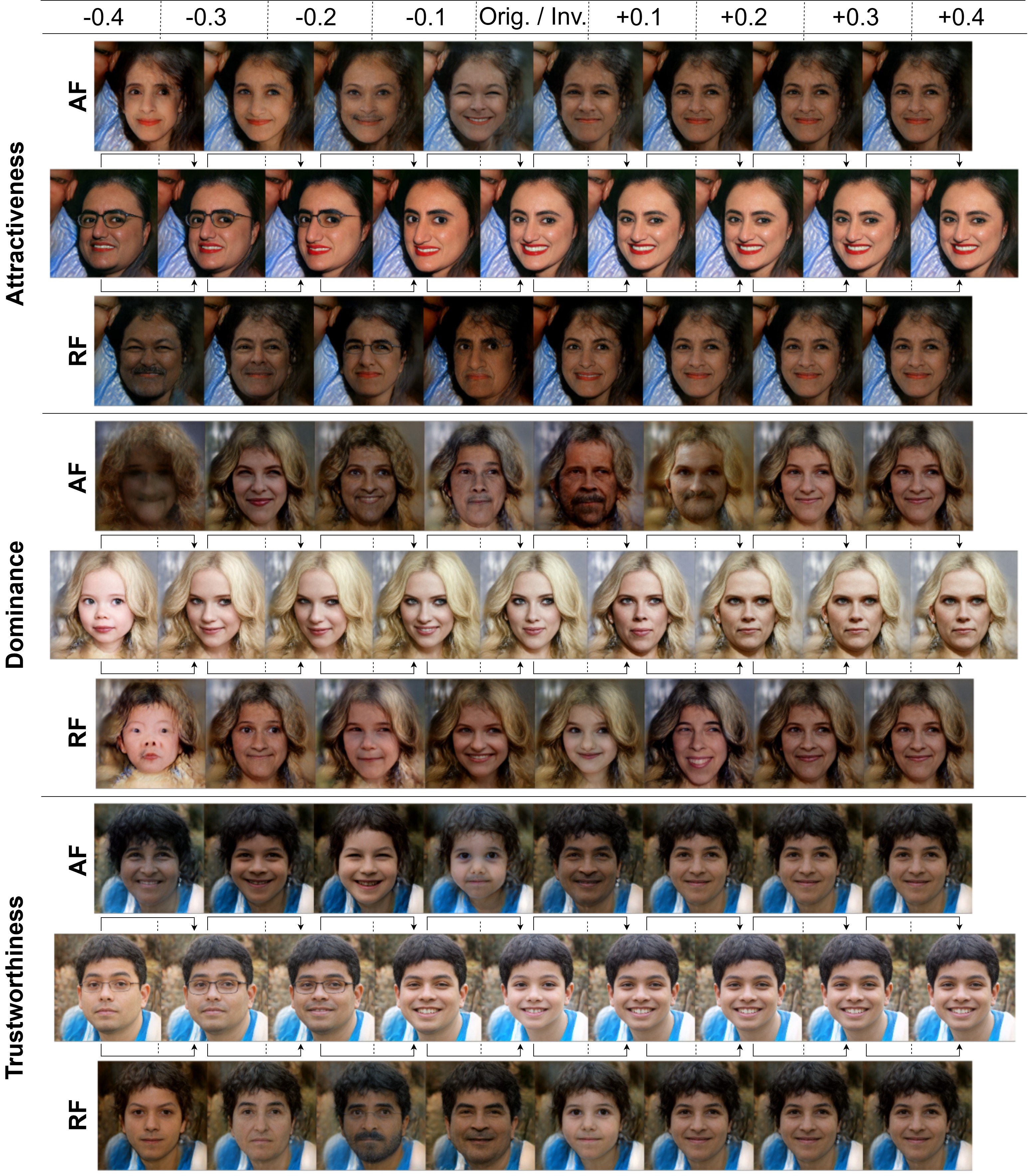}
    \vspace{0.15cm}
    \caption{Middle row shows \textit{face transformations} for an example face image. Top and bottom row present feature changes performed by the pipeline to obtain the desired subjective attribute score. AF corresponds to \textit{added features} and RF stands for \textit{removed features}.
    Consider the transition from -0.4 to -0.3; AF shows the addition of age along with an increase in smile. Alternatively, RF shows the removal of baby-like nose (visible nostrils), wider eyes, rounded eyebrows, and fuller lips. Transitions from left to right represent \textit{increasing elicited scores}. The reverse can be obtained by adding RF and removing the AF features. Absence of significant changes results in similar visualizations for AF and RF.}
\label{fig:fp_performance}
\end{figure*}

\newpage

\section{Additional Qualitative Results (Section~\ref{section:overall_results})}
\label{section:additional_qualitative_results}
\vspace{-0.2cm}
\subsection{CelebAMask-HQ ``ID Loss'' training setup, Trustworthiness}
\vspace{-0.4cm}
\begin{figure}[htbp]
    \centering
    \includegraphics[width=15cm]{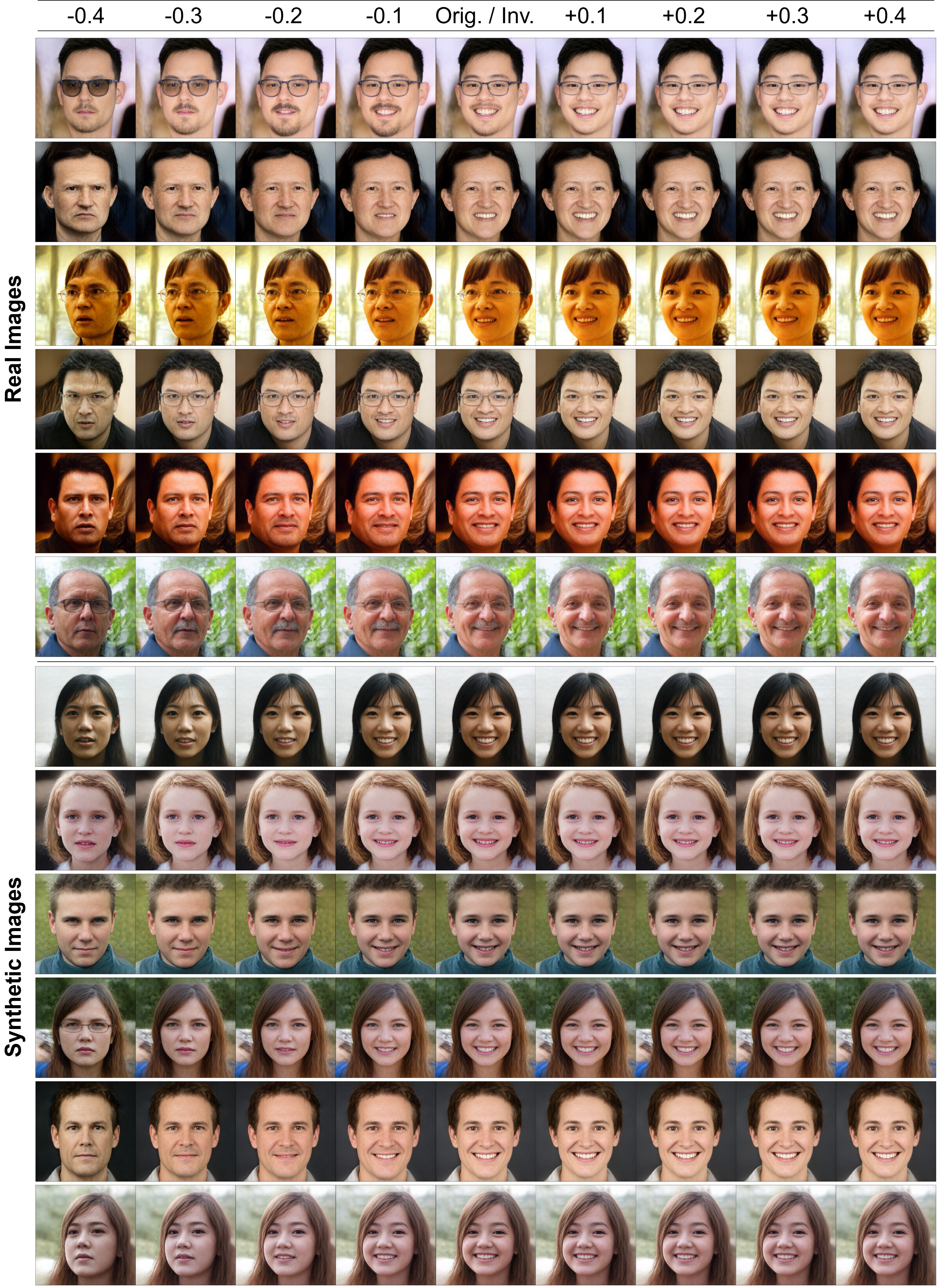}
    \caption{Qualitative results: Real images are from FFHQ evaluative set, Synthetic images are from OMI.}
\label{fig:fp_performance_1}
\end{figure}

\newpage
\subsection{CelebAMask-HQ ``ID Loss'' training setup, Dominance}

\begin{figure}[htbp]
    \centering
    \includegraphics[width=15cm]{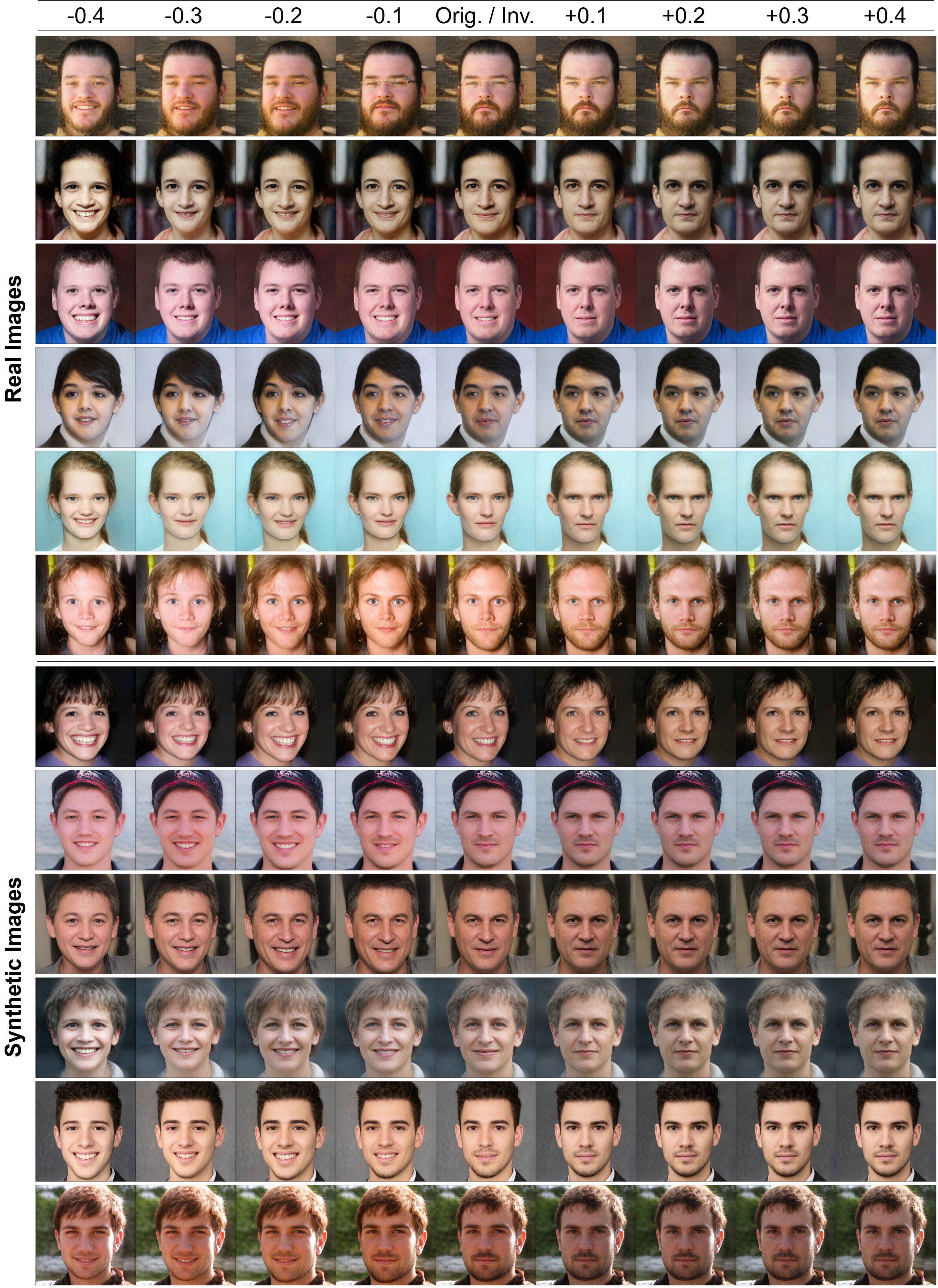}
    \caption{Qualitative results: Real images are from FFHQ evaluative set, Synthetic images are from OMI.}
\label{fig:fp_performance_2}
\end{figure}

\newpage
\subsection{CelebAMask-HQ ``ID Loss'' training setup, Attractiveness}

\begin{figure}[htbp]
    \centering
    \includegraphics[width=15cm]{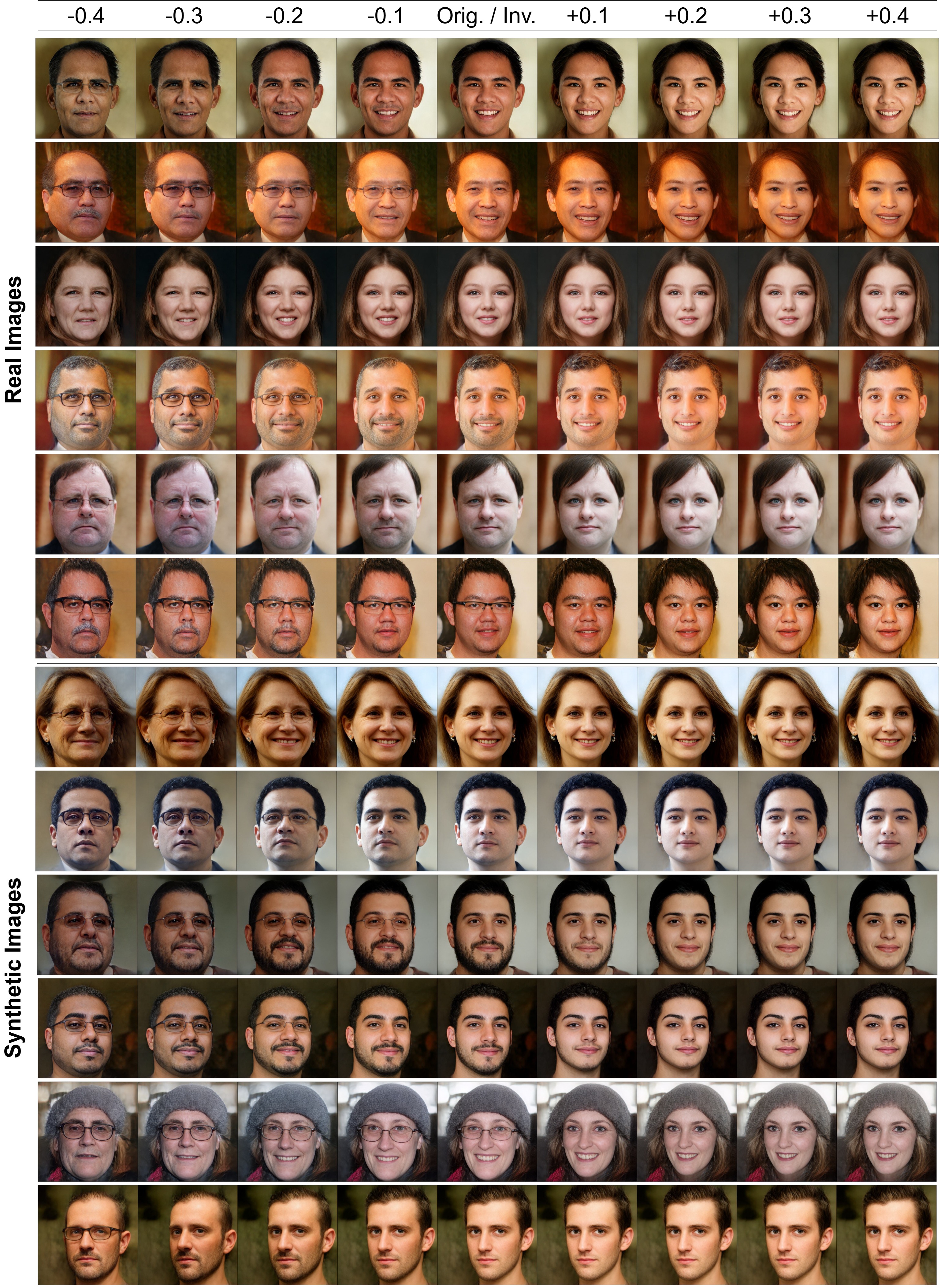}
    \caption{Qualitative results: Real images are from FFHQ evaluative set, Synthetic images are from OMI.}
\label{fig:fp_performance_3}
\vspace{-0.5cm}
\end{figure}

\newpage
\section{Human Study Setup Details}
\label{section:human_study_details}

\subsection{Edit Consistency with Human Perception (Section \ref{section:edit_consistency_with_human_perception})}
\label{section:consistency_human_perception}

We conduct a human study with 54 participants, recruited from the United States through Prolific, to verify whether our pipeline outputs images are consistent with human perceptions. Each participant was provided with the study instructions, after which they signed an IRB approved consent form before responding to study questions. Stimuli for this experiment included 100 images each selected from the FFHQ and Synthetic evaluation sets (StyleGAN-ADA generated images), for each attribute. For each selected image, we pair the original image with a transformed version; these were transformations with one of the considered attribute score changes i.e., \{-0.2, -0.1, +0.1, +0.2\}. Each target change was equally represented in the full set. We further shuffled the order of the original and transformed versions, such that the original image was placed on either side with equal proportion. 

\begin{figure}[htbp]
    \centering
    \includegraphics[width=13cm]{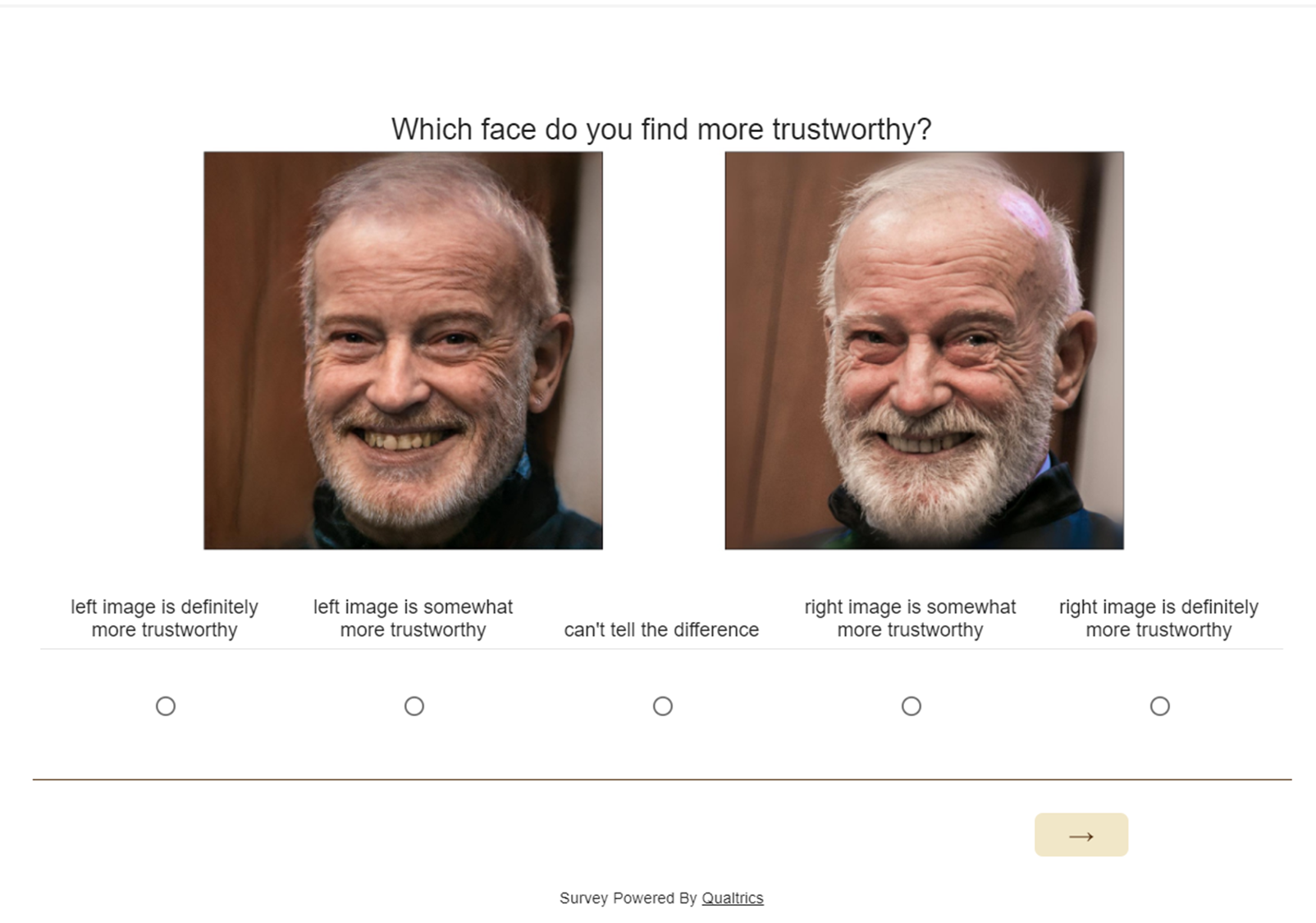}
    \caption{Human study setup presented during the instructional video --- provided to the participants at the beginning of the study. }
    \label{fig:human_study_example}
\end{figure}

Each participant was shown 30-35 randomly selected image pairs, and were compensated \$1.00, with an additional \$0.20 given as bonus if they answered within the first 12 seconds on average. This was done to incentivize speed to capture responses closer to first impressions. The total study time was for 10 minutes. For each image pair, response to the question: ``Which face do you find more trustworthy / dominant / attractive?" was collected from 3 participants. They selected one of five options mapped to a Likert scale \cite{likert1932technique, lucey2010extended, langner2010presentation}, which also took into account their level of confidence. The options ranged from left image definitely being more trustworthy to right definitely being more trustworthy. The middle option included can't tell the difference --- this meant that there was no visible perceptual change in the edit from the perspective of the rater (Fig. \ref{fig:human_study_example}).

\subsection{Identity Preservation Ability (Section \ref{section:identity_preservation})}
\label{section:human_study_id_preservation}

We conduct an additional study to verify the identity preservation ability of the pipeline, with a similar setup as discussed in Section \ref{section:consistency_human_perception}. 27 participants are recruited from the United States through Prolific and are provided with relevant instructions, after which they sign an IRB approved consent form for answering the study questions. Stimuli for this experiment included 100 images combined selected from the FFHQ and Synthetic evaluation sets (StyleGAN-ADA generated images), for each attribute. For each selected image, we pair adjacent images across the transformation spectrum $[-0.2, 0.2]$. For example, $-0.2$ is paired with $-0.1$, $-0.1$ is paired with the \textit{original}, and so on. This created 4 possible pairs in a given transformation spectrum for an identity --- each possibility was equally represented in the full set.

\begin{figure}[htbp]
    \centering
    \vspace{0.25cm}
    \includegraphics[width=13cm]{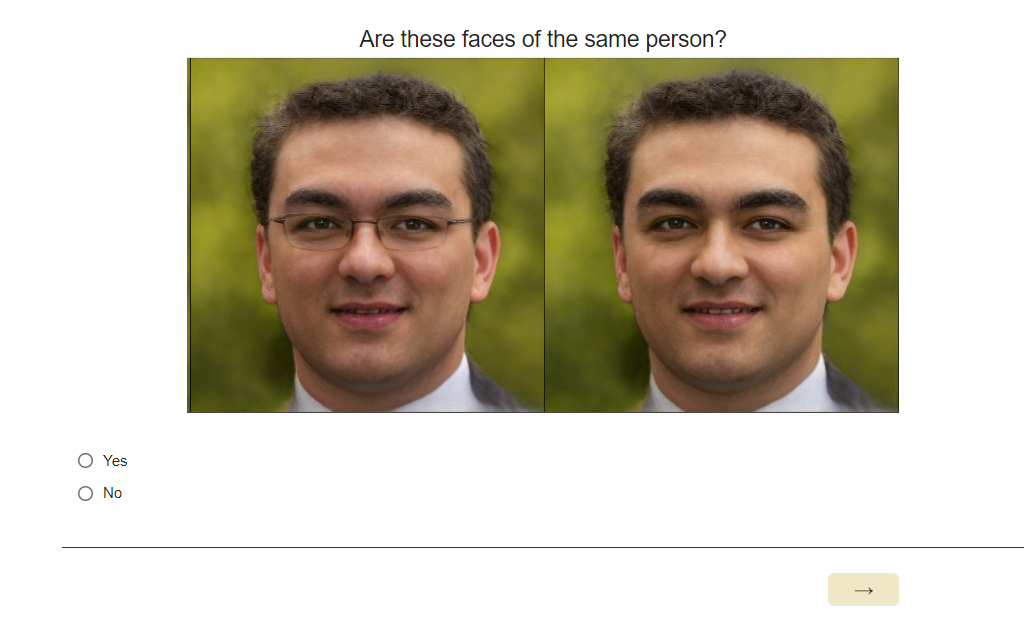}
    \caption{Human study setup to verify identity retention using the proposed method. }
    \label{fig:human_study_setup_identity}
\end{figure}

Each participant was shown 30-35 randomly selected image pairs, and were compensated \$1.00, with an additional \$0.20 given as bonus if they answered within the first 12 seconds on average. This was done to incentivize speed to capture responses closer to first impressions. The total study time was for 10 minutes. For each image pair, response to the question: ``Are these faces of the same person?" was collected from 3 participants. They selected from either a \textit{Yes} or a \textit{No} (Fig. \ref{fig:human_study_setup_identity}). If the majority (atleast 2 out of 3) participants responded with a \textit{Yes} for an image, we consider that image to retain the identity during the transformation.

\newpage
\section{Identity Preservation Ability: Qualitative Results (Section \ref{section:identity_preservation})}
\label{section:omi_comparison}

\begin{figure}[htbp]
    \centering
    \vspace{0.25cm}
    \includegraphics[width=13cm]{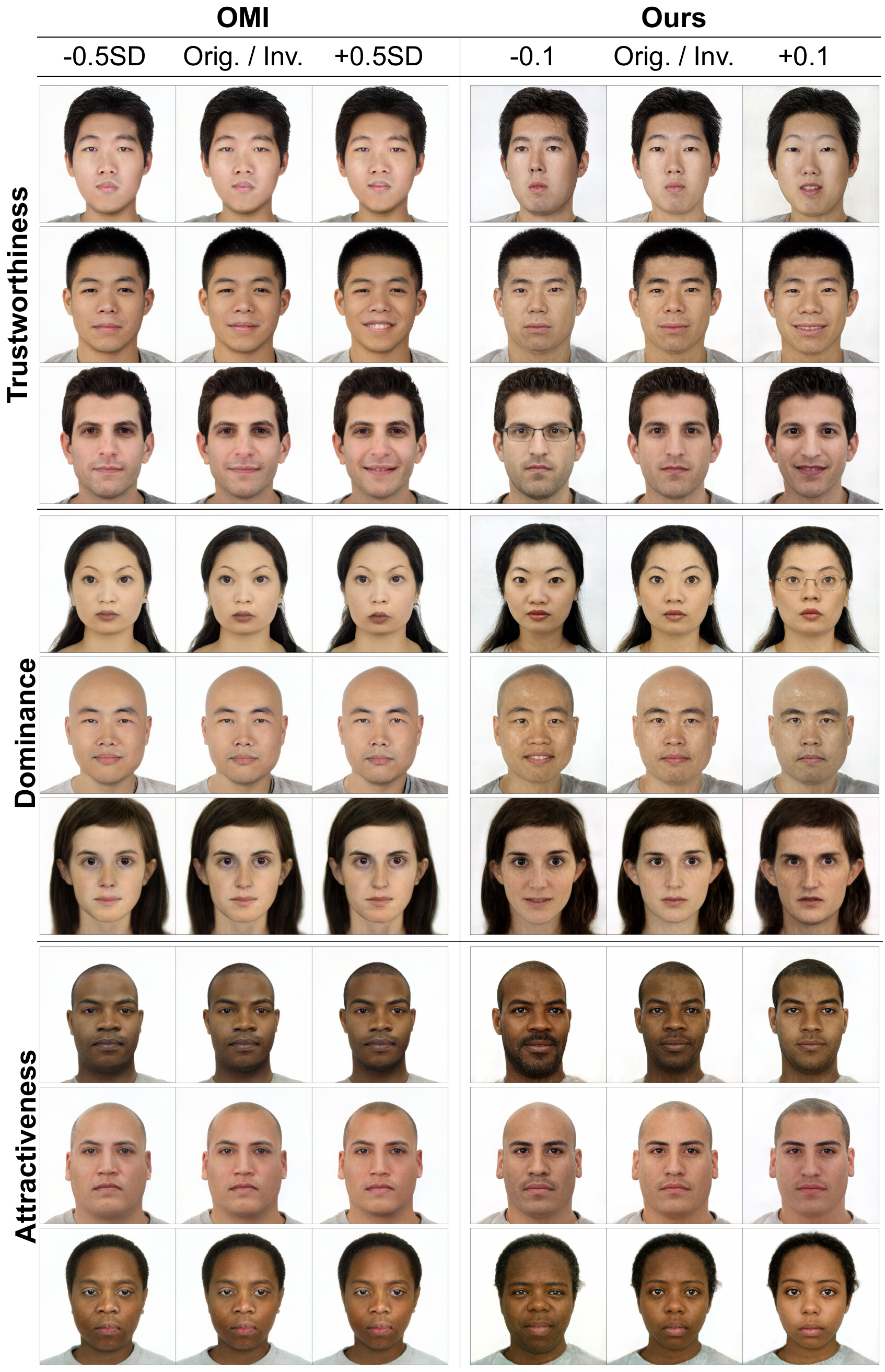}
    \caption{Comparison of our transformation method with \cite{peterson2022omi}. Their edits are in the range \textpm 0.5 SD of the mean rating of an attribute, while ours are between \textpm 0.1. }
    \label{fig:omi_compare}
    \vspace{-0.5cm}
\end{figure}

\newpage
\section{Impact of Training Data on Edits: Feature Diversity (Section \ref{section:overall_results})}

\begin{figure*}[htbp]
    \centering
    \vspace{0.25cm}
    \includegraphics[width=17cm]{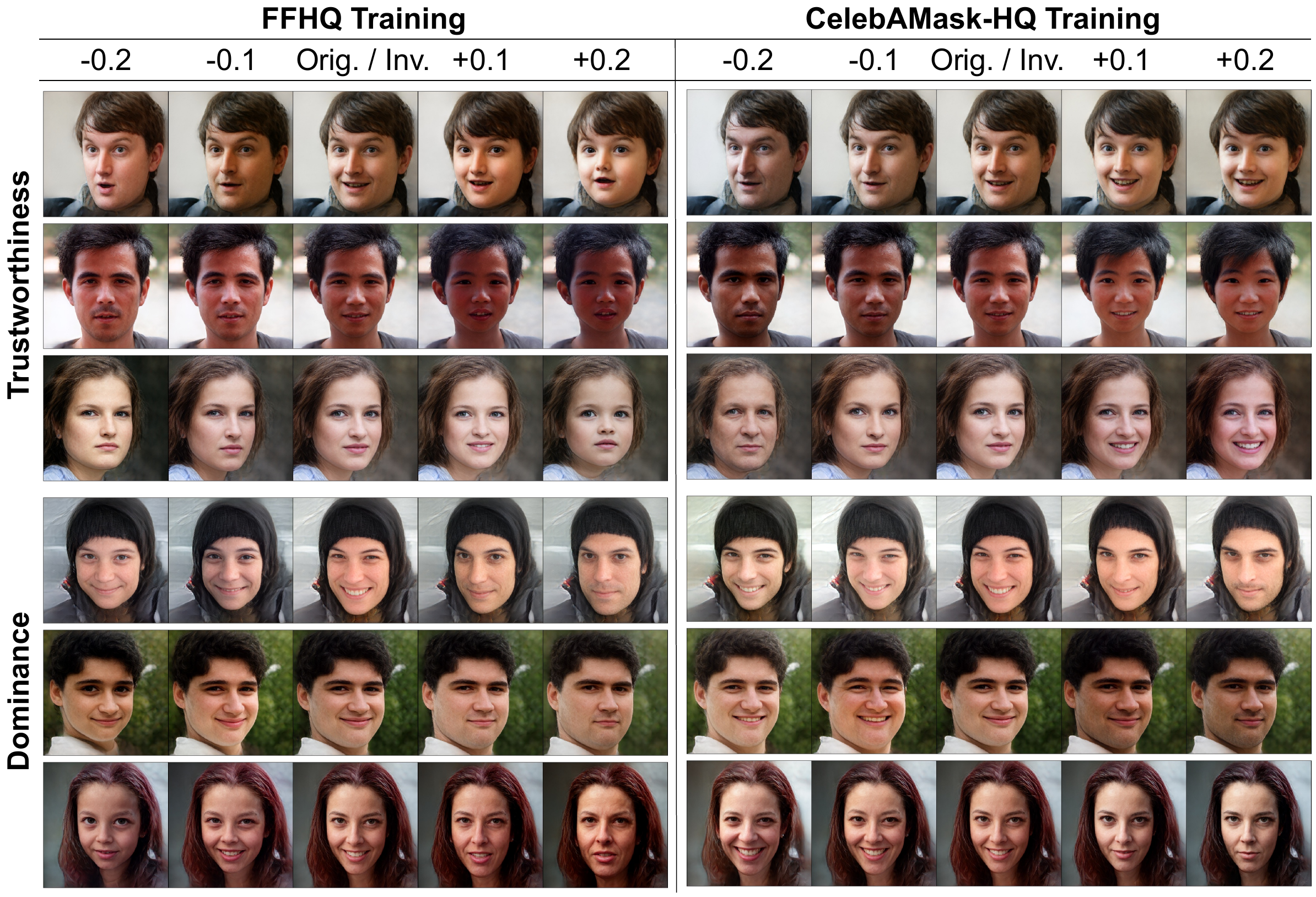}
    \caption{Training data image feature diversity affecting the transformations. We observe apparent differences in edits when comparing both sides, like \textit{decrease in age} during the reduction in dominance perception or increase in trustworthiness perception, or even subtle changes like difference in hairstyles or the subject's eye color.}
\label{fig:ffhq_vs_celebamask_performance_supp}
\end{figure*}

The images show semantically continuous transformations on either side of the original. This is true for both \textit{real} and \textit{synthetic} evaluation sets\footnote{Additional qualitative results are presented in Section~\ref{section:additional_qualitative_results}.}. The results show correlations of subjective attributes with combinations of various objective edits. One common observation is related to \textit{age} --- while the identity's age $\uparrow$ with $\uparrow$ in \textit{dominance}, $\uparrow$ in \textit{trustworthiness} \& \textit{attractiveness} is associated with $\downarrow$ in age. This leads to the interpretation that older individuals tend to be perceived as more dominant~\cite{batres2015influence}, while younger individuals are perceived as more trustworthy and attractive~\cite{sutherland2022understanding}. A few other observations follow the same correlation; $\uparrow$ in dominance is seemingly associated with $\uparrow$ in facial hair, wrinkles, and angry facial expressions, while an $\uparrow$ in the other attributes correspond to $\downarrow$ in the same objective edits. The opposite is true for facial edits involving a \textit{smile} --- $\uparrow$ in smile is associated with $\uparrow$ in trustworthiness \& attractiveness, and $\downarrow$ in dominance. The pipeline learns these correlations using face images annotated via human-like subjective prediction models, extending from crowd perception data used to train them.

\newpage

\section{Synthetic Data to Improve First Impression Prediction: An Application (Main Paper Section VI)}
\label{section:improve_fi_prediction_supp}

\vspace{-0.3cm}
\begin{table*}[htbp]
\begin{center}
\begin{small}
\caption{Original image datasets with color image stimuli and corresponding subjective attribute labels. Note that we train impression prediction models on the average of crowd-sourced annotations for an image.}
\vspace{-0.2cm}
\label{table:original_datasets}
\begin{sc}
\renewcommand{\arraystretch}{1.3}
\begin{tabular}{ >{\raggedright\arraybackslash}p{1.35cm} 
>{\raggedright\arraybackslash}p{1cm}  >{\raggedright\arraybackslash}p{0.5cm} 
>{\raggedright\arraybackslash}p{2.15cm}  >{\raggedright\arraybackslash}p{4.1cm}  >{\raggedright\arraybackslash}p{4.05cm} 
>{\raggedright\arraybackslash}p{0.75cm}}
 \hline
{\bfseries Dataset} & {\bfseries Images} & {\bfseries Lab} & {\bfseries Race} & {\bfseries Expression} & {\bfseries Subjective Attribute} & {\bfseries Rate}\\
 \hline
CFD~\cite{ma2015chicago, ma2021chicago} & 827 & Yes & Asian, Latino, Black, White & Neutral, happy (open/close mouth), angry, fearful & attractive, trustworthy, dominant (597 images) & 1-7 \\
 \hline
 UAF~\cite{bainbridge2013intrinsic} & 2222 & No & White, Black, Asian, Hispanic & - & attractive, trustworthy & 1-9 \\
 \hline
 SCUT~\cite{liang2018scut} & 5500 & No & Asian, Caucasian & - & attractiveness & 1-5 \\
 \hline
 OMI~\cite{peterson2022omi} & 1004 & No & \footnotesize{\textit{SYNTHETIC}} & - & trustworthy, attractive, dominant & [0,1] \\
 \hline
\end{tabular}
\end{sc}
\end{small}
\end{center}
\vspace{-0.5cm}
\end{table*}

\end{appendix}

\end{document}